\documentclass[letterpaper]{article} 
\usepackage{aaai2026}  
\usepackage{times}  
\usepackage{helvet}  
\usepackage{courier}  
\usepackage[hyphens]{url}  
\usepackage{graphicx} 
\usepackage{natbib}  
\usepackage{caption} 
\usepackage{algorithm}
\usepackage{algorithmic}

\usepackage{newfloat}
\usepackage{listings}
\DeclareCaptionStyle{ruled}{labelfont=normalfont,labelsep=colon,strut=off} 
\floatstyle{ruled}
\newfloat{listing}{tb}{lst}{}
\floatname{listing}{Listing}
\usepackage{latexsym}
\usepackage{amssymb}
\usepackage{amsmath}
\usepackage{amsthm}
\usepackage{booktabs}
\usepackage{enumitem}

\usepackage{multirow}
\usepackage{mathtools}  
\usepackage{bm}         
\usepackage{booktabs}
\usepackage{mathrsfs}   
\usepackage{physics}    
\usepackage{tabularx}

\usepackage{times}
\usepackage{url}
\usepackage[utf8]{inputenc}
\usepackage{amsmath}
\usepackage{amsfonts}
\usepackage{booktabs}
\usepackage{appendix}

\title{SPIRIT: Short-Term Prediction of Solar IRradIance for Transfer Learning Using Foundation Models}
\author {
    Aditya Mishra\equalcontrib\textsuperscript{\rm 1},
    T Ravindra\equalcontrib\textsuperscript{\rm 1},
    Srinivasan Iyengar\textsuperscript{\rm 2},
    Shivkumar Kalyanaraman\textsuperscript{\rm 3}, 
    Ponnurangam Kumaraguru\textsuperscript{\rm 1}
}
\affiliations{
    \textsuperscript{\rm 1}International Institute of Information Technology, Hyderabad\\
    \textsuperscript{\rm 2}Microsoft Corporation India\\
    \textsuperscript{\rm 3}Anusandhan National Research Foundation\\
    \{aditya.mishra, t.ravindra\}@students.iiit.ac.in,
    pk.guru@iiit.ac.in,
    sriyengar@microsoft.com,
    ceo@anrf.gov.in
}

\begin{document}
\maketitle

\begin{abstract}
Traditional solar forecasting models are based on several years of site-specific historical irradiance data, often spanning five or more years, which are unavailable for newer photovoltaic farms. As renewable energy is highly intermittent, building accurate solar irradiance forecasting systems that are data-efficient is essential for efficient grid management and enabling the ongoing proliferation of solar energy, which is crucial to achieve the United Nations' net zero goals. In this work, we propose SPIRIT, a novel framework leveraging foundation models for solar irradiance forecasting, making it applicable to newer solar installations. Our approach outperforms state-of-the-art models in zero-shot transfer learning by upto 70\%, enabling effective performance at new locations without relying on any past data. Further improvements in performance are achieved through fine-tuning, as more location-specific data becomes available. These findings are supported by statistical significance, further validating our approach. By dramatically reducing the forecasting setup timeline, SPIRIT accelerates solar farm deployment in all potential global sites, most of which lack historical data, thereby democratizing access to clean energy and enabling participation in the renewable energy transition.
\end{abstract}

\begin{links}
    \link{Code}{https://github.com/surya-ravindra06/SPIRIT-Official}
\end{links}

\begin{figure*}[h]
    \centering
    \includegraphics[width=\textwidth]{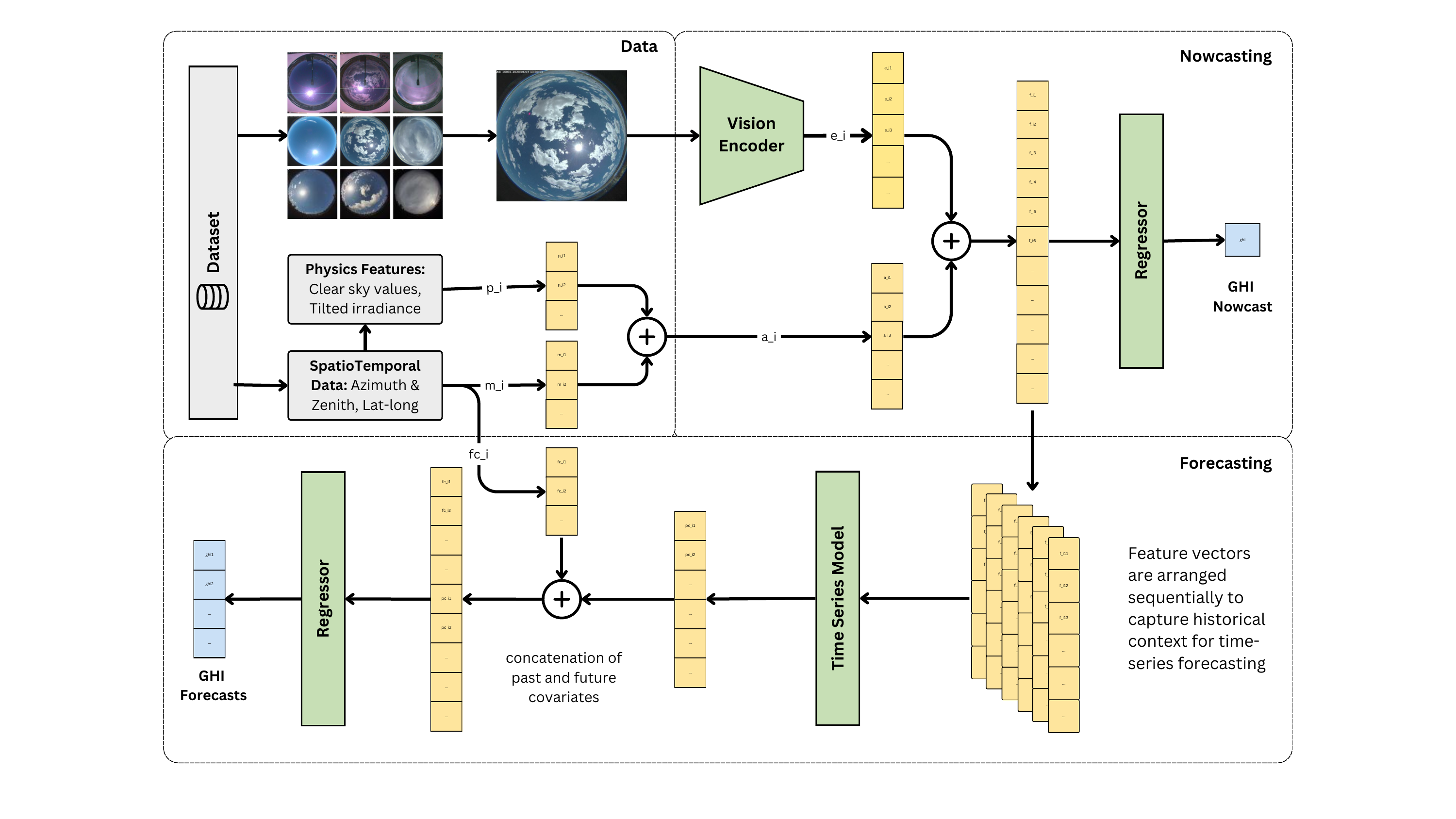}
    \caption{
    Illustration of our system: A vision encoder (top-left) extracts embeddings from a sky camera image sampled from a diverse set spanning multiple locations and setups. Physics-inspired features are derived and integrated with auxiliary values, then merged with the image embedding (top-middle) into a unified representation. For nowcasting (right), a regressor predicts Global Horizontal Irradiance from this feature vector. For forecasting (bottom), a time-series model processes past feature vectors to create a context embedding, which is concatenated with a future covariate vector constructed from known future values to form the final latent representation. A regressor then maps this representation to future GHI values (bottom-right).
}  
    \label{fig:crown_jewel}
\end{figure*}

\section{Introduction}
The proliferation of solar energy is paramount for electrification and the global energy transition to meet the Net Zero commitments of the United Nations~\cite{sadhukhan2022net}. Solar energy is notable for its potential to reduce carbon emissions~\cite{sen2008solar}. Expanding the solar energy infrastructure is crucial to mitigate the effects of climate change~\cite{bashir2021enabling} and meet the energy demands arising from sectors such as data centers~\cite{agarwal2021redesigning}, transportation~\cite{lee2016shared}, and buildings~\cite{iyengar2017cloud}.

Unlike conventional sources, solar energy presents unique challenges due to its intermittency from daily and seasonal sunlight variations, significantly impacting grid stability~\cite{grid_management}. High solar penetration creates the "duck curve" phenomenon~\cite{iyengar2016analyzing}, where misalignment between solar production and peak demand complicates grid management. Despite increasing storage capacity, electricity grids function primarily as \textit{just-in-time} systems requiring balanced energy supply and demand~\cite{joskow2012creating}. To maintain grid reliability, operators face deviation penalties for unplanned energy contributions~\cite{yang2020penalty}, necessitating accurate short-term solar predictions for efficient grid operation~\cite{iyengar2014solarcast}.

Existing approaches for short-term forecasting use sky cameras, a fish-eye lens positioned to look directly towards the zenith, which require extensive site-specific data to train models~\cite{hendrikx2024_short_term_forecast, wacv2022}. While these methods have demonstrated high accuracy, they rely on multi-year training datasets. With global solar photovoltaic (PV) capacity projected to expand from 1 terawatt in 2022 to 10 terawatts by 2030~\cite{isa2023}, 90\% of the solar farms worldwide will have negligible data to train custom models from scratch. Thus, lack of site-specific data underscores the need for novel approaches.

Vision foundation models have revolutionized numerous computer vision tasks such as feature extraction and object detection, enabling improved accuracy through zero-shot and few-shot approaches with limited data~\cite{google_vit, object_detection_foundation_models, vits_gt_cnns1}. Concurrently, physics-inspired feature engineering has enhanced model performance by integrating domain-specific knowledge, yielding more accurate and interpretable predictions~\cite{ompusunggu2021physics,erdmann2020physics}. Our research hypothesis investigates: \textit{Can we leverage state-of-the-art vision foundation models and physics-inspired features, along with transfer learning strategies, to reduce the dependence on site-specific sky camera data?}

To address these challenges, we introduce SPIRIT, a novel solar irradiance forecasting approach with inductive bias toward enhanced generalizability. Our contributions include:

(1) Development of an innovative system leveraging foundation models, physics-informed features and future covariates that eliminates site-specific training requirements while facilitating adaptation across diverse transfer learning scenarios. Our system's flexibility enables seamless integration of future vision model advancements.

(2) Motivated by real-world deployment constraints, we demonstrate that SPIRIT can rapidly scale to new solar plant locations without prior sky camera data, significantly accelerating integration into operational workflows.

\section{Related Work}
Traditional solar forecasting methods primarily utilize Numerical Weather Prediction models and satellite imagery~\cite{markovics2022comparison}. These approaches often lack sufficient spatial and temporal resolution for accurate short-term forecasts. NWP models typically operate at kilometer-scale grids with updates every few hours, failing to capture rapid cloud cover changes affecting solar irradiance~\cite{kostylev2011solar}. Recent years have seen various time series forecasting approaches for solar prediction, yet these generally function at multi-hour to day-ahead timescales, rendering them inadequate for detecting short-term solar generation variations from transient factors like cloud cover~\cite{iyengar2014solarcast}.

Sky cameras have gained prominence due to their potential to enhance short-term solar power prediction~\cite{wacv2022,yuhao_transfer_learning}. These fish-eye lenses capture wide-angle sky images that provide crucial data on cloud cover and movement, key determinants of solar irradiance forecasting~\cite{dev2019estimating}. Recent research has focused on utilizing sky cameras to overcome prior limitations. \citet{wacv2022} and \citet{hammond2024} demonstrated sky cameras' efficacy in developing high-precision short-term solar forecasting models. Their studies employed extensive site-specific data collected across multiple years, yielding models with significantly improved forecast accuracy compared to conventional methods.

\citet{talha2019} developed a deep learning framework combining sky-camera imagery with meteorological data for solar irradiance prediction. Their approach integrates a CNN and temporal LSTM for forecasts up to four hours ahead, demonstrating that auxiliary data (temperature, wind speed, relative humidity) enhances prediction generalization when trained on 10 years of data. \citet{wacv2022} proposed a transformer-based architecture incorporating a clear sky model to estimate residual irradiance, achieving superior accuracy compared to CNN-LSTM methods using 10 years of training data. Both studies emphasize the efficacy of integrating sky imagery with auxiliary data for precise solar forecasting. However, these approaches face data availability constraints, particularly significant as global solar PV capacity expands from 1 TW (2022) to a projected 10 TW by 2030, leaving most solar installations around the world with insufficient historical data for custom model development.

Limited availability of site-specific data presents a critical challenge for advancing solar forecasting. Despite significant improvements in short-term predictions through sky cameras and auxiliary data integration, these methods face scalability constraints due to insufficient historical data at most solar installations. Transfer learning emerges as a promising solution, enabling knowledge leveraging from pre-trained models and adaptation of learned representations across diverse datasets and locations. Recent research~\cite{yuhao_transfer_learning, paletta_transfer_learning} demonstrates that multi-dataset fusion training and data augmentation methods yield models with superior performance, highlighting cross-dataset knowledge transfer benefits in solar forecasting.

\section{Design and Implementation}
\subsection{Key Concepts}
\textbf{Nowcasting and Forecasting:}
Nowcasting, the prediction of solar power generation over very short time horizons, typically ranges from a few minutes to a few hours \cite{nowcasting_defn}. In contrast, short-term forecasting extends the prediction horizon to cover periods from one hour to 24 hours \cite{forecasting_defn}. Methods developed to provide forecasts utilize various data sources, such as satellite data \cite{nowcasting_satellite, nowcasting_defn}, weather station observations \cite{nowcasting_defn}, and sky camera images \cite{wacv2022, talha2019}. Nowcasting and short-term forecasting are indispensable for managing the intermittency of solar power, allowing grid operators to perform better scheduling, dispatching, and balancing of energy resources \cite{aouidad2024machine}.

\textbf{Sky Camera:} 
Sky cameras enhance nowcasting and short-term forecasting by capturing sky images with fish-eye lenses, providing detailed cloud movement, and sun position data. These images enable algorithms to track cloud dynamics and predict their trajectories, essential for estimating solar irradiance \cite{saraswat2023sky, dev2019estimating}. Offering a low-latency alternative to weather satellites, sky cameras facilitate real-time monitoring. However, variations in camera setup and quality affect image appearance, as shown in Figure 4 in the Appendix. As a key tool in solar forecasting, sky cameras contribute to more reliable energy predictions \cite{rajagukguk2021deep}. Further details are provided in Appendix C.
 
\textbf{Solar Irradiance} necessitates distinguishing between three fundamental measurements. Direct Normal Irradiance (DNI) quantifies solar radiation received per unit area perpendicular to the sun's rays without atmospheric scattering. Diffuse Horizontal Irradiance (DHI) measures scattered radiation reaching a horizontal surface from all sky directions, becoming crucial during cloud cover. \textit{Global Horizontal Irradiance} (GHI), the total radiation on a horizontal surface, encompasses both of these components and is mathematically expressed as:
\begin{equation}
    GHI = DNI \times \cos(\theta) + DHI
\end{equation}
where $\theta$ represents the zenith angle, defined as the angle between the direction of incoming solar radiation and the vertical.

GHI is the most commonly used irradiance measure in solar energy applications, as it directly influences photovoltaic (PV) panel performance and solar power generation, making it the primary focus of research in irradiance forecasting. Henceforth, unless explicitly stated otherwise, any mention of irradiance or solar irradiance
refers specifically to Global Horizontal Irradiance.

\textbf{Photovoltaic Power Output} refers to the electricity generated by solar panels from incoming solar radiation. While it is primarily driven by GHI \cite{ghi_pv_linear}, factors like temperature, and system losses also play a role. Under stable conditions, the relationship between GHI and PV output is roughly linear \cite{pv_output}. Since PV output is a more actionable metric for grid management and energy planning, predicting it directly is often more desirable.

\subsection{Physics-inspired Feature Engineering}
\label{subsec:physics_inspired}

\textbf{Clear sky models}~\cite{ineichen1, clearsky1} are mathematical models that estimate the solar irradiance at a given location under cloud-free conditions, serving as a representation of the maximum possible radiation reaching the Earth's surface. These models leverage fundamental atmospheric physics and employ mathematical formulations based on solar geometry, atmospheric transmittance, and radiative transfer~\cite{clearsky1}. The Ineichen model~\cite{ineichen1} requires inputs such as latitude, longitude, time, and date, which are readily available. This allows clear sky irradiance values to be readily computed and incorporated into our model as features, providing a reference for expected irradiance levels in the absence of cloud interference.

\textbf{Physics behind solar irradiance:}
Solar irradiance varies significantly based on site-specific parameters like panel orientation and solar positioning. The critical angle of incidence ($\theta$), representing the angle between incoming solar rays and panel normal, is calculated as $\cos(\theta)=\cos(\theta_{z}) \cdot \cos(\beta) + \sin(\theta_{z}) \cdot \sin(\beta) \cdot \cos(\gamma - \alpha)$, where $\theta_{z}$ and $\gamma$ denote solar zenith and azimuth angles, while $\beta$ and $\alpha$ represent panel tilt and azimuth angles, respectively.

These physics-inspired features enable the creation of future covariates through their deterministic calculation capabilities independent of temporal constraints, thereby enhancing time-series information for predictive modeling.

\subsection{Nowcasting Architecture}
\label{subsec:nowcasting_architecture}

We propose an architecture that encodes sky images into vector representations, which are augmented with auxiliary data and physics-based features to capture GHI information that is then effectively extracted by a regression model. Let \( \mathcal{X} \) be the set of sky camera images, and \( \mathcal{D} \) be the dataset, defined as \( \mathcal{D} = \{(X_i, \mathbf{A}_i, y_i)\}_{i=1}^N \), where \( X_i \in \mathcal{X} \) represents the \(i\)-th sky image, \( \mathbf{A}_i \in \mathbb{R}^k \) corresponds to the auxiliary features such as azimuth and zenith angles of the Sun, and \( y_i \in \mathbb{R}^+ \) are the corresponding solar irradiance measurements.

An encoder function \( E: \mathcal{X} \rightarrow \mathbb{R}^d \) is defined that assigns a \( d \)-dimensional embedding vector to each image \( X \in \mathcal{X} \):

\[
\mathbf{Z} = E(X), \quad \mathbf{Z} \in \mathbb{R}^d
\]

To leverage domain knowledge in solar power prediction, we introduce a set of additional features, \( \mathbf{P} \), derived from the auxiliary measurements \( \mathbf{A} \). These features incorporate established solar engineering principles, such as clear sky irradiance, and panel tilt and orientation, as defined in Subsection \ref{subsec:physics_inspired}. The feature vector is given by \( \mathbf{P} \in \mathbb{R}^p \) where \( p \) represents the number of physics-based features extracted from the auxiliary data.

The final feature representation \( \mathbf{f} \in \mathbb{R}^{d+k+p} \) is constructed by concatenating the image embedding \( \mathbf{Z} \), raw auxiliary measurements \( \mathbf{A} \), and the physics-based features \( \mathbf{P} \):
\[
\mathbf{f} = \mathbf{Z} \oplus \mathbf{A} \oplus \mathbf{P}
\]
where \( \oplus \) denotes concatenation. This combined representation leverages data-driven, visual and domain-specific features, providing a comprehensive characterization of each sample \( (X_i, \mathbf{A}_i, y_i) \in \mathcal{D} \).

A regression function \( R_\omega: \mathbb{R}^{d+k+p} \rightarrow \mathbb{R}^+ \), parameterized by weights \( \omega \), is defined such that:

\[
\hat{y} = R_\omega(\mathbf{f}) = R_\omega(E(X) \oplus \mathbf{A} \oplus \mathbf{P})
\]

Nowcasting loss function \( \mathcal{L}_{nowcast}(\omega) \) is defined as the average of the individual regression losses for each sample, where each individual loss measures the discrepancy between the predicted \( \hat{y}_i = R_\omega(\mathbf{f}_i) \) and the true value \( y_i \) across all $N$ samples in the dataset:

\[
\mathcal{L}_{nowcast}(\omega) = \frac{1}{N} \sum_{i=1}^N \mathcal{L}(R_\omega(\mathbf{f}_i), y_i)
\]

where \( \mathcal{L}(R_\omega(\mathbf{f}_i), y_i) \) is the regression loss for the \(i\)-th sample. To learn the optimal parameters \( \omega^* \), we minimize \( \mathcal{L}_{nowcast}(\omega) \) using gradient-based methods.

\subsection{Forecasting Architecture}

Our forecasting architecture processes sequences of sky images to predict GHI across multiple future intervals. Each image is encoded using the embedding and augmentation approach from Section~\ref{subsec:nowcasting_architecture}, creating a foundation for temporal analysis. A sequence of \( T \) images \( X_{1:T} = \{ X_1, X_2, \dots, X_T \} \) along with their corresponding auxiliary features \( \mathbf{A}_{1:T} = \{ \mathbf{A}_1, \mathbf{A}_2, \dots, \mathbf{A}_T \} \), where each \( \mathbf{A}_t \in \mathbb{R}^k \) represents the auxiliary feature vector at time \( t \), is given.

An encoder function \( E \) generates the vector representation \( \mathbf{Z}_t = E(X_t) \in \mathbb{R}^d \) for each image at time step \( t = 1, 2, \dots, T \). The physics-based features \( \mathbf{P}_t \) are derived from auxiliary measurements \( \mathbf{A}_t \). The final feature vectors \( \mathbf{f}_t \in \mathbb{R}^{d+k+p} \) are obtained by concatenating the image embedding, auxiliary data, and physics-based features:
\[
\mathbf{f}_t = \mathbf{Z}_t \oplus \mathbf{A}_t \oplus \mathbf{P}_t
\]

where \( \oplus \) denotes concatenation, providing a comprehensive characterization of each sample \( (X_t, \mathbf{A}_t, y_t) \in \mathcal{D} \). Thus, the set of feature vectors for each timestamp over the sequence of \( T \) time steps is given by \( \mathbf{F_{1:T}} = \{ \mathbf{f}_1, \mathbf{f}_2, \dots, \mathbf{f}_T \} \).


A time-series model \( \mathcal{M} \) is used to encode the observed sequence \( \mathbf{F_{1:T}} \) into a latent vector \( \mathbf{L} \in \mathbb{R}^m \), which captures the full context of the input data series while retaining its temporal patterns and dependencies:

\[
\mathbf{L} = \mathcal{M}(\mathbf{F_{1:T}}) \in \mathbb{R}^m
\]

To integrate known future information, derived from the spatiotemporal context of time and location, future covariate vectors \( \mathbf{C}_{T+\tau_i} \in \mathbb{R}^{q} \) are constructed for each forecast time \( T + \tau_i \). The full covariate vector \( \mathbf{C} \in \mathbb{R}^{q \cdot H} \) is then formed by concatenating these individual representations across all $H$ forecast horizons:

\newcommand{\medoplus}{\mathop{\mathchoice
  {\displaystyle\oplus}
  {\textstyle\oplus}
  {\scriptstyle\oplus}
  {\scriptscriptstyle\oplus}}}

\[
\mathbf{C} = \medoplus_{i=1}^{H} \mathbf{C}_{T+\tau_i}, \quad \mathbf{C}_{T+\tau_i} \in \mathbb{R}^{q}
\]


We concatenate the future covariate vector \( \mathbf{C} \) with the latent representation of the past time steps \( \mathbf{L} \), forming the final vector \( \mathbf{h} \) that encompasses all relevant information: \( \mathbf{h} = \mathbf{L} \oplus \mathbf{C} \). This ensures that both past contextual information as well as known future data contribute to the forecasting process. A regression function \( R_\omega: \mathbb{R}^{m+q \cdot H} \to \mathbb{R}^H \), parameterized by \( \omega \), is applied to the vector \( \mathbf{h} \in \mathbb{R}^{m+q \cdot H}\) to generate the corresponding predicted GHI values. The regressor outputs a vector \( \hat{\mathbf{y}} \in \mathbb{R}^H \) of predicted GHI values for the forecast time intervals \( T + \tau_1, T + \tau_2, \dots, T + \tau_H \):

\[
\hat{\mathbf{y}} = R_\omega(\mathbf{h}) = \left[ \hat{y}_{T+\tau_1}, \hat{y}_{T+\tau_2}, \dots, \hat{y}_{T+\tau_H} \right] \in \mathbb{R}^H
\]

where each \( \hat{y}_{i} \) corresponds to the irradiance forecast for the time interval \( T + \tau_i \), and the vector \( \hat{\mathbf{y}} \) represents the full set of predicted irradiance values across all forecast intervals \( T + \tau_1, T + \tau_2, \dots, T + \tau_H \). 

Forecasting loss function \( \mathcal{L}_{forecast}(\omega) \) is defined as the mean of the individual regression losses computed over all forecast intervals \( T + \tau_j \) for each sample \( i \) across $N$ samples and $H$ forecast horizons. The total loss is given by:

\[
\mathcal{L}_{forecast}(\omega) = \frac{1}{N \cdot H} \sum_{i=1}^N \sum_{j=1}^H \mathcal{L}(\hat{y}^{(i)}_{T+\tau_j}, y^{(i)}_{T+\tau_j})
\]

where \( \mathcal{L}(\hat{y}^{(i)}_{T+\tau_j}, y^{(i)}_{T+\tau_j}) \)  is the individual regression loss for the forecast interval \( T + \tau_j \) for sample \( i \) . To learn the optimal parameters \( \omega^* \), we minimize \( \mathcal{L}_{forecast}(\omega) \) using gradient-based optimization methods. The complete architecture is illustrated in Figure \ref{fig:crown_jewel}.

\textbf{The Significance of Generalized Encoders:}
A key distinction of our approach is that in prior work \cite{wacv2022, prior_methods_site_specific, talha2019}, the encoder \( E \) is a vision model typically trained on data from a specific location and camera setup. Furthermore, studies aiming for generalizability typically rely on training models using a fusion of solar datasets from multiple locations \cite{yuhao_transfer_learning, site_specific_trained_transfer_learning}. In contrast, we argue, and later demonstrate, that leveraging a foundation model, a highly generalizable feature extractor, provides a more robust \( E \) function. A foundation model not only matches the performance of site-specific encoders at a given location with a particular setup but also demonstrates an unparalleled advantage in generalizing across diverse locations and camera setups.

\subsection{Implementation}
Our implementation leverages the pre-trained Google ViT~\cite{google_vit} for image embedding generation, deliberately excluding all meteorological sensor data to significantly reduce sensor dependence. Instead, we incorporate only the zenith and azimuth angles and, physics-based features (clear sky irradiance values, panel tilt and orientation). These carefully selected image embeddings are concatenated with the auxiliary vector to form robust feature representations that train an XGBoost regressor for nowcasting. For forecasting, we extend these feature representations across a one-hour context window (six consecutive timestamps) and process them through a transformer encoder~\cite{vaswani2017attention}. The resulting temporal representation is fused with future covariates from the auxiliary and physics features before passing through MLP layers to predict solar irradiance at 1-4 hour horizons, using clear-sky GHI as residuals before the final regressor. This represents one concrete realization of our system, with comprehensive ablations in Appendix~A for understanding the contributions of each component. More details on hyperparameters and deployment efficiency are provided in Appendix~F.
       
\section{Evaluation Methodology}
\subsection{Datasets and Performance Metrics}
We evaluate our methods using three publicly available datasets: TSI880~\cite{tsi_dataset}, ASI16~\cite{tsi_dataset}, and SKIPP'D~\cite{SKIPPD_dataset}. The TSI880 and ASI16 datasets, both collected from the NREL Solar Radiation Research Laboratory in Golden, Colorado, provide sky images captured every 10 minutes along with corresponding GHI values and auxiliary data such as air temperature and relative humidity and only differ in camera setup and sensors, with the ASI16 dataset capturing higher-resolution images. The SKIPP'D dataset, collected from Stanford University, consists of raw sky images captured every minute and PV power output data, prioritizing finer temporal granularity at the expense of image quality. For more details, refer to Appendix~B.

We utilize the TSI880 and ASI16 datasets to investigate the impact of camera setup at the same location. To explore location and task shifts, we use the SKIPP'D dataset to evaluate the performance of models trained on GHI data in predicting PV power output. The SKIPP'D dataset features lower-resolution images and lacks meteorological data, thereby presenting a more challenging task by limiting the contextual information typically leveraged by prior models~\cite{wacv2022, talha2019}. To ensure the models learn from higher-quality, information-rich datasets, we train exclusively on the TSI and ASI datasets while evaluation is done across all the datasets, including the more challenging SKIPP'D, allowing us to assess how well the models generalize to lower-quality data and increased domain shifts.

We assess the effectiveness of the predicted values using the normalized Mean Absolute Percentage error (nMAP):

\begin{equation}
\text{nMAP} = \frac{1}{N} \sum_{i=1}^{N} \frac{|y_i - \hat{y}_i|}{\frac{1}{N} \sum_{i=1}^{N} y_i} \times 100 
\end{equation}

where \( y_i \) represents the actual value and \( \hat{y}_i \) represents the predicted value for the \( i \)-th sample, with \( i \in \{1, \dots, N\} \). It is commonly used for solar irradiance prediction as the normalization ensures that models can be assessed uniformly on different datasets, avoiding biased assessments due to scale differences. Comprehensive analysis on metrics is in Appendix~I.

\subsection{Baselines}

We benchmark against \citet{wacv2022}, the current state-of-the-art, who achieve superior performance by training a vision transformer~\cite{google_vit} and temporal transformer~\cite{vaswani2017attention} on 10 years of site-specific data~\cite{tsi_dataset, wacv2022, talha2019} with sensor-dependent auxiliary values. For equitable evaluation, we reproduced their architecture and conducted experiments under identical conditions for both their model and SPIRIT. More details on baselines are in Appendix~G.

\begin{table}[t]
  \centering
  \renewcommand{\arraystretch}{1.2}
  \begin{tabular}{@{}c c c c@{}}
    \hline
    \textbf{Train} & \textbf{Test} & \textbf{SPIRIT} & \textbf{\citeauthor{wacv2022}} \\
    \hline
    \multirow{3}{*}{TSI} & ASI & \textbf{27.17} \small{{(-62.49)}} & 89.66 \\
                          & SKIPP'D & \textbf{35.94} \small{{(-60.94)}} & 96.43 \\
                          & TSI & 9.04 \small{{(+0.08)}} & \textbf{8.96} \\
    \cline{1-4}
    \multirow{3}{*}{ASI} & TSI & \textbf{28.86} \small{{(-46.65)}} & 75.51 \\
                          & SKIPP'D & \textbf{32.98} \small{{(-57.69)}} & 90.67 \\
                          & ASI & 9.08 \small{{(+0.95)}} & \textbf{8.13} \\
    \hline
  \end{tabular}
  \caption{
  Nowcasting performance across multiple datasets: SPIRIT and \citet{wacv2022} model trained on one dataset for a year are evaluated with nMAP error (lower is better) both in a zero-shot setting and on the same dataset, with testing on TSI 2021, ASI 2021, and SKIPP'D 2017. Performance is comparable when tested in the training setup, but our model demonstrates significantly better zero-shot performance when in a new setup or a new location.}
  \label{tab:zeroshot_nowcast}
\end{table}

\begin{table}[t]
  \centering
  \setlength{\tabcolsep}{2pt}
  \renewcommand{\arraystretch}{1.2} 
  \begin{tabular}{c c c c c}
    \hline
    \textbf{Interval} & \textbf{Train} & \textbf{Test} & \textbf{SPIRIT} & \textbf{\citeauthor{wacv2022}} \\
    \hline
    \multirow{5}{*}{1hr} & \multirow{2}{*}{TSI} & ASI & \textbf{29.99}\hspace{1mm}\small{{(-5.75)}} & 35.74 \\
                          & & SKIPP'D & \textbf{32.93}\hspace{1mm}\small{{(-5.95)}} & 38.88 \\
                          & & TSI & \textbf{18.96}\hspace{1mm}\small{{(-1.00)}} & 19.96 \\
                          \cline{2-5}
                          & \multirow{2}{*}{ASI} & TSI & \textbf{26.85}\hspace{1mm}\small{{(-2.19)}} & 29.04 \\
                          & & SKIPP'D & \textbf{27.33}\hspace{1mm}\small{{(-14.35)}} & 41.68 \\
                          & & ASI & 19.23\hspace{1mm}\small{{(+0.02)}} & \textbf{19.21} \\
    \hline
    \multirow{5}{*}{2hr} & \multirow{2}{*}{TSI} & ASI & \textbf{31.71}\hspace{1mm}\small{{(-5.89)}} & 37.60 \\
                          & & SKIPP'D & \textbf{29.01}\hspace{1mm}\small{{(-14.80)}} & 43.81 \\
                          & & TSI & \textbf{21.77}\hspace{1mm}\small{{(-0.87)}} & 22.64 \\
                          \cline{2-5}
                          & \multirow{2}{*}{ASI} & TSI & \textbf{28.64}\hspace{1mm}\small{{(-1.01)}} & 30.65 \\
                          & & SKIPP'D & \textbf{26.29}\hspace{1mm}\small{{(-21.63)}} & 47.92 \\
                          & & ASI & \textbf{21.51}\hspace{1mm}\small{{(-0.47)}} & 21.98 \\
    \hline
    \multirow{5}{*}{3hr} & \multirow{2}{*}{TSI} & ASI & \textbf{34.41}\hspace{1mm}\small{{(-3.36)}} & 37.77 \\
                          & & SKIPP'D & \textbf{30.26}\hspace{1mm}\small{{(-17.10)}} & 47.36 \\
                          & & TSI & \textbf{25.46}\hspace{1mm}\small{{(-0.84)}} & 26.30 \\
                          \cline{2-5}
                          & \multirow{2}{*}{ASI} & TSI & \textbf{31.65}\hspace{1mm}\small{{(-1.5)}} & 33.15 \\
                          & & SKIPP'D & \textbf{30.26}\hspace{1mm}\small{{(-22.89)}} & 53.15 \\
                          & & ASI & \textbf{24.78}\hspace{1mm}\small{{(-0.89)}} & 25.67 \\
    \hline
    \multirow{5}{*}{4hr} & \multirow{2}{*}{TSI} & ASI & \textbf{38.00}\hspace{1mm}\small{{(-1.58)}} & 39.58 \\
                          & & SKIPP'D & \textbf{34.63}\hspace{1mm}\small{{(-17.15)}} & 51.78 \\
                          & & TSI & \textbf{29.89}\hspace{1mm}\small{{(-1.69)}} & 31.58 \\
                          \cline{2-5}
                          & \multirow{2}{*}{ASI} & TSI & \textbf{35.86}\small\hspace{1mm}{{(-0.99)}} & 36.85 \\
                          & & SKIPP'D & \textbf{36.97}\hspace{1mm}\small{{(-13.20)}} & 50.17 \\
                          & & ASI & \textbf{29.29}\hspace{1mm}\small{{(-1.73)}} & 31.02 \\
    \hline
  \end{tabular}
  \caption{
  Forecasting performance across multiple datasets and forecast intervals: SPIRIT and \citet{wacv2022} model trained on one dataset are evaluated with nMAP error both in a zero-shot setting and on the same dataset, with testing on TSI 2021, ASI 2021, and SKIPP'D 2017 across four forecast intervals: 1hr, 2hr, 3hr, and 4hr.}
  \label{tab:zeroshot_forecast}
\end{table}

\subsection{Zero-shot Transfer Learning}
\label{subsec:zero_shot_transfer_learning}
To evaluate the zero-shot generalization performance of our models, we analyze two distinct transfer learning scenarios. First, we examine intra-location generalization where models are trained and tested at identical geographic locations but with differing camera configurations. Though environmental conditions remain consistent, variations in camera setup alter the spatial mapping of critical features within images. Models trained on specific camera setups learn to associate image regions with key elements (sun position, cloud formations, atmospheric conditions) that influence predictions. When these spatial relationships shift due to altered camera configurations, model performance deteriorates. We evaluate this scenario by training on TSI and testing on ASI (both from Golden, Colorado), and vice versa.

Second, we investigate cross-location and cross-task generalization, training models in one geographic location and testing in another with different environmental and sensor characteristics. Specifically, we train on TSI and ASI datasets and evaluate on SKIPP'D (from Stanford University, California), shifting from GHI prediction to PV power output forecasting. Given the nearly linear correlation between GHI and PV output \cite{ghi_pv_linear}, this constitutes a valid heterogeneous transfer learning scenario. Model outputs are normalized to address scale differences between GHI and PV measurements, as detailed in Appendix~F.2. For comprehensive evaluation, we conduct experiments for both nowcasting and forecasting tasks, training on one year and testing on another to account for seasonal variations. Tables~\ref{tab:zeroshot_nowcast} and~\ref{tab:zeroshot_forecast} report nMAP errors for nowcasting and forecasting respectively, comparing SPIRIT against the state-of-the-art in both zero-shot transfer learning and traditional setting (same location and setup).

\begin{figure}[t]
    \centering
    \includegraphics[width=150 pt]{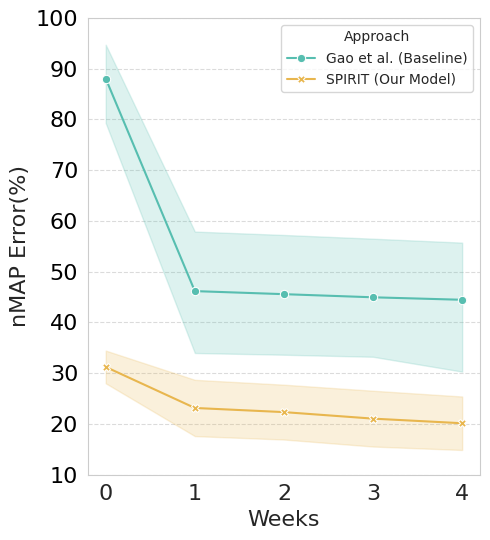}
    \caption{
    Mean nMAP performance of SPIRIT and \citet{wacv2022} across varying fine-tuning data sizes in weeks. Solid lines denote average accuracy and shaded regions show 95\% confidence intervals across multiple runs, including cross-setup (TSI to ASI and vice versa) and cross-location (TSI or ASI to SKIPP’D) transfer scenarios.}
    \label{fig:finetune_nowcast}
\end{figure}

\begin{figure*}[t]
    \centering
    \includegraphics[width=\textwidth]{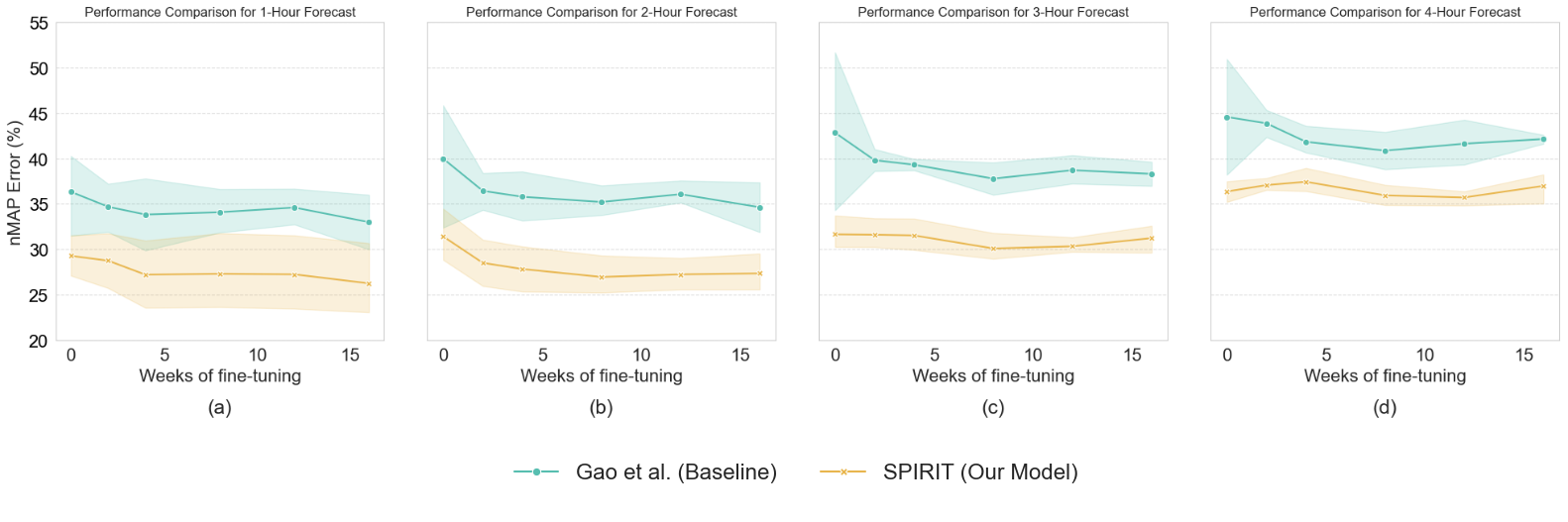}
    \caption{
Forecasting performance of SPIRIT and \citet{wacv2022} using nMAP error across different forecast intervals. Subfigures (a), (b), (c), and (d) correspond to 1-hour, 2-hour, 3-hour, and 4-hour forecasting, respectively. The solid lines represent the average performance, with varying fine-tuning data sizes (in weeks). The shaded regions denote the 95\% confidence interval, illustrating the variability across multiple runs, including training on one dataset and finetuning and testing on another, as well as utilizing randomized seasonal sampling. SPIRIT exhibits consistently better performance and low variance compared to the baseline, particularly with severely limited data, demonstrating its ability to maintain stability over the baseline's.}

    \label{fig:finetune_forecast}
\end{figure*}

\subsection{Fine-tuning with Limited Data}
Following our zero-shot transfer learning experiments, we examine SPIRIT's adaptability through fine-tuning with limited target domain data. This scenario replicates real-world deployment constraints where extensive data collection is impractical and models must adapt rapidly with minimal supervision. We evaluate both intra-location and cross-location/cross-task adaptation scenarios as described in Subsection~\ref{subsec:zero_shot_transfer_learning}. For both experimental configurations, we incrementally increase the amount of target domain data used for fine-tuning, one to four weeks for nowcasting and two to sixteen weeks for forecasting, with testing performance on the remaining annual data. The longer timeframe for forecasting accounts for the reduced sample density due to temporal consistency constraints, highlighted in Appendix~B.3, and forecasting's fundamentally higher complexity relative to nowcasting.

We employ a selective fine-tuning strategy wherein only the regressor layers (Figure~\ref{fig:crown_jewel}) are updated while freezing the remainder of the temporal encoder. This preserves the pre-trained feature extractors that capture generalizable spatiotemporal patterns, an essential part of our core hypothesis. Selective fine-tuning of terminal layers not only accelerates adaptation but also mitigates overfitting on limited target data~\cite{yuhao_transfer_learning, uses_freezing_for_tl, transfer_learning_strategies_comparison}. Detailed rationale for this choice is provided in Appendix~D.3. The nowcasting and forecasting performance metrics are presented in Figures~\ref{fig:finetune_nowcast} and~\ref{fig:finetune_forecast}, respectively.

\section{Results}
\subsection{Zero-shot Transfer Learning}

Tables~\ref{tab:zeroshot_nowcast} and~\ref{tab:zeroshot_forecast} present the results for zero-shot transfer learning, demonstrating that our model consistently outperforms the state-of-the-art baseline across both cross-location and cross-setup scenarios in nowcasting and forecasting tasks. In nowcasting, SPIRIT exhibits remarkable adaptability when transitioning between camera setups within the same location (TSI to ASI and vice versa) or across different geographical contexts (TSI and ASI to SKIPP'D), outperforming the baseline by up to 70\%. Similarly, in forecasting tasks, our approach achieves up to 45\% improvement when transferring across locations (from TSI/ASI to SKIPP'D). These substantial performance gains, particularly in challenging cross-location and cross-task settings, underscore our model's exceptional generalizability and robustness. Furthermore, even in the traditional setup where models are trained and tested on the same dataset and setup, SPIRIT demonstrates enhanced or comparable forecasting performance, further emphasizing its effectiveness across diverse deployment conditions. Additional results for our system with different foundation model variants are provided in Appendix~E, confirming that the observed improvements stem from SPIRIT's design.

\subsection{Fine-tuning with Limited Data}

Our experimental analysis merges cross-setup evaluations (training on ASI, finetuning and subsequently testing on TSI, and vice versa) and cross-location scenarios (shifting from ASI or TSI to SKIPP'D) to ensure a sufficient number of experimental instances for the robust confidence interval plots in Figures~\ref{fig:finetune_nowcast} and~\ref{fig:finetune_forecast} for nowcasting and forecasting respectively. In the simpler nowcasting task, both models exhibit rapid improvement within the first week. However, the baseline reaches performance saturation around 45\%, while SPIRIT continues to reduce its error, achieving a significant improvement, dropping below 20\% within four weeks. 

In forecasting, SPIRIT consistently outperforms the baseline, demonstrating notably lower variance, particularly in data-limited settings in the first 2 weeks. This underscores SPIRIT's superior stability and reliability, with its nMAP error remaining consistently below that of the baseline. In contrast, the baseline exhibits higher variance, indicating greater inconsistency and confusion in its performance when limited data is available. Both models experience a typical performance decline as the forecasting horizon extends from 1-hour to 4-hour forecasts, driven by the increased uncertainty over longer time horizons. Nonetheless, SPIRIT’s consistently lower variance and sustained performance highlight its robustness and its ability to adapt more effectively to challenging conditions. The transition from a zero-shot configuration to fine-tuning results in noticeable performance improvements; however, the gains diminish after approximately eight weeks of fine-tuning, suggesting that extended fine-tuning beyond this period yields only marginal additional benefits. Detailed results are in Appendix~D.


\section{Conclusion}
This work addresses a critical challenge in solar irradiance forecasting: generalizing models to novel geographic locations without historical data. SPIRIT leverages pre-trained models and physics-informed features for future covariates, enabling robust performance in data-scarce environments while facilitating post-deployment fine-tuning as site-specific observations accumulate. SPIRIT's modular architecture accommodates emerging vision models, ensuring its methodological relevance while its scalability accelerates the deployment of solar farms in remote and emerging markets, while also enhancing the reliability, cost-effectiveness, and accessibility of solar energy generation to facilitate the global transition toward renewable energy systems.

\section{Future Work and Limitations}
A key limitation is the geographical restriction of datasets to North America due to limited availability of publicly accessible data from other regions, while Southern Hemisphere solar dynamics require further investigation. Future work will expand dataset diversity to enhance generalizability. Additionally, foundation models introduce computational overhead and inference latency. Subsequent research will focus on efficiency optimization for deployment on resource-constrained edge devices.

\section*{Acknowledgements}
We would like to thank the PreCog Lab at IIIT Hyderabad for their valuable guidance and discussions throughout this work. In particular, we thank Sreeram Vennam, Jain Hemang Ashok, Vaishnavi Shivkumar, Akshit Sinha, Anish Joishy and Vamshi Krishna Bonagiri for their detailed feedback and thoughtful suggestions during multiple stages of the project. We also thank Harpreet Singh, Quentin Paletta, Talha Ahmad Siddiqui, and Tejas Cavale for their valuable inputs and constructive perspectives. Finally, we thank the anonymous reviewers for their insightful feedback. The views expressed are those of the authors and do not necessarily reflect the official policies or positions of the supporting organizations.

\bibliography{aaai2026}

@article{aouidad2024machine,
  title={Machine learning-based short-term solar power forecasting: a comparison between regression and classification approaches using extensive Australian dataset},
  author={Aouidad, Hichem Idris and Bouhelal, Abdelhamid},
  journal={Sustainable Energy Research},
  volume={11},
  number={1},
  pages={28},
  year={2024},
  publisher={Springer}
}

@article{saraswat2023sky,
  title={Sky Image Classification Based Solar Power Prediction Using CNN.},
  author={Saraswat, Rahul and Jhanwar, Deepak and Gupta, Manish},
  journal={Traitement du Signal},
  volume={40},
  number={4},
  year={2023}
}

@article{dev2019estimating,
  title={Estimating solar irradiance using sky imagers},
  author={Dev, Soumyabrata and Savoy, Florian M and Lee, Yee Hui and Winkler, Stefan},
  journal={Atmospheric Measurement Techniques},
  volume={12},
  number={10},
  pages={5417--5429},
  year={2019},
  publisher={Copernicus Publications G{\"o}ttingen, Germany}
}

@article{rajagukguk2021deep,
  title={A deep learning model to forecast solar irradiance using a sky camera},
  author={Rajagukguk, Rial A and Kamil, Raihan and Lee, Hyun-Jin},
  journal={Applied Sciences},
  volume={11},
  number={11},
  pages={5049},
  year={2021},
  publisher={MDPI}
}

@inproceedings{bashir2021enabling,
  title={Enabling sustainable clouds: The case for virtualizing the energy system},
  author={Bashir, Noman and Guo, Tian and Hajiesmaili, Mohammad and Irwin, David and Shenoy, Prashant and Sitaraman, Ramesh and Souza, Abel and Wierman, Adam},
  booktitle={Proceedings of the ACM Symposium on Cloud Computing},
  pages={350--358},
  year={2021}
}

@inproceedings{agarwal2021redesigning,
  title={Redesigning data centers for renewable energy},
  author={Agarwal, Anup and Sun, Jinghan and Noghabi, Shadi and Iyengar, Srinivasan and Badam, Anirudh and Chandra, Ranveer and Seshan, Srinivasan and Kalyanaraman, Shivkumar},
  booktitle={Proceedings of the 20th ACM Workshop on Hot Topics in Networks},
  pages={45--52},
  year={2021}
}

@inproceedings{iyengar2016analyzing,
  title={Analyzing energy usage on a city-scale using utility smart meters},
  author={Iyengar, Srinivasan and Lee, Stephen and Irwin, David and Shenoy, Prashant},
  booktitle={Proceedings of the 3rd ACM International Conference on Systems for Energy-Efficient Built Environments},
  pages={51--60},
  year={2016}
}

@book{sen2008solar,
  title={Solar energy fundamentals and modeling techniques: atmosphere, environment, climate change and renewable energy},
  author={Sen, Zekai},
  year={2008},
  publisher={Springer Science \& Business Media}
}

@article{sadhukhan2022net,
  title={Net zero electricity systems in global economies by life cycle assessment (LCA) considering ecosystem, health, monetization, and soil CO2 sequestration impacts},
  author={Sadhukhan, Jhuma},
  journal={Renewable Energy},
  volume={184},
  pages={960--974},
  year={2022},
  publisher={Elsevier}
}

@article{iyengar2017cloud,
  title={A cloud-based black-box solar predictor for smart homes},
  author={Iyengar, Srinivasan and Sharma, Navin and Irwin, David and Shenoy, Prashant and Ramamritham, Krithi},
  journal={ACM Transactions on Cyber-Physical Systems},
  volume={1},
  number={4},
  pages={1--24},
  year={2017},
  publisher={ACM New York, NY, USA}
}

@article{yang2020penalty,
  title={A penalty scheme for mitigating uninstructed deviation of generation outputs from variable renewables in a distribution market},
  author={Yang, Jiajia and Dong, Zhao Yang and Wen, Fushuan and Chen, Qixin and Luo, Fengji and Liu, Weijia and Zhan, Junpeng},
  journal={IEEE Transactions on Smart Grid},
  volume={11},
  number={5},
  pages={4056--4069},
  year={2020},
  publisher={IEEE}
}

@inproceedings{lee2016shared,
  title={Shared solar-powered EV charging stations: Feasibility and benefits},
  author={Lee, Stephen and Iyengar, Srinivasan and Irwin, David and Shenoy, Prashant},
  booktitle={2016 Seventh International Green and Sustainable Computing Conference (IGSC)},
  pages={1--8},
  year={2016},
  organization={IEEE}
}

@techreport{isa2023,
title = {World Solar Market Report 2023},
institution = {International Solar Alliance},
year = {2023},
author = {ISA}
}

@article{joskow2012creating,
  title={Creating a smarter US electricity grid},
  author={Joskow, Paul L},
  journal={Journal of Economic Perspectives},
  volume={26},
  number={1},
  pages={29--48},
  year={2012},
  publisher={American Economic Association}
}

@inproceedings{iyengar2014solarcast,
  title={SolarCast: a cloud-based black box solar predictor for smart homes},
  author={Iyengar, Srinivasan and Sharma, Navin and Irwin, David and Shenoy, Prashant and Ramamritham, Krithi},
  booktitle={Proceedings of the 1st ACM Conference on Embedded Systems for Energy-Efficient Buildings},
  pages={40--49},
  year={2014}
}

@misc{hammond2024,
    title={Short-Term Solar Irradiance Forecasting Under Data Transmission Constraints}, 
    author={Joshua Edward Hammond and Ricardo A. Lara Orozco and Michael Baldea and Brian A. Korgel},
    year={2024},
    eprint={2403.12873},
    archivePrefix={arXiv}
    }

@article{markovics2022comparison,
  title={Comparison of machine learning methods for photovoltaic power forecasting based on numerical weather prediction},
  author={Markovics, D{\'a}vid and Mayer, Martin J{\'a}nos},
  journal={Renewable and Sustainable Energy Reviews},
  volume={161},
  pages={112364},
  year={2022},
  publisher={Elsevier}
}

@inproceedings{kostylev2011solar,
  title={Solar power forecasting performance--towards industry standards},
  author={Kostylev, Vladimir and Pavlovski, Alexandre and others},
  booktitle={1st international workshop on the integration of solar power into power systems, Aarhus, Denmark},
  pages={1--8},
  year={2011},
  organization={Energynautics GmbH M{\"u}hlstra{\ss}e Langen, Germany}
}

@inproceedings{ompusunggu2021physics,
  title={Physics-Inspired Feature Engineering for Condition Monitoring of Alternating Current-Powered Solenoid-Operated Valves},
  author={Ompusunggu, Agusmian Partogi and Hostens, Erik},
  booktitle={International Conference on Maintenance, Condition Monitoring and Diagnostics},
  pages={139--151},
  year={2021},
  organization={Springer}
}

@inproceedings{erdmann2020physics,
  title={Physics inspired feature engineering with Lorentz Boost Networks},
  author={Erdmann, M and Geiser, E and Rath, Y and Rieger, M},
  booktitle={Journal of Physics: Conference Series},
  volume={1525},
  number={1},
  pages={012107},
  year={2020},
  organization={IOP Publishing}
}

@INPROCEEDINGS{wacv2022,
  author={Gao, Huiyu and Liu, Miaomiao},
  booktitle={2022 IEEE/CVF Winter Conference on Applications of Computer Vision (WACV)}, 
  title={Short-term Solar Irradiance Prediction from Sky Images with a Clear Sky Model}, 
  year={2022},
  volume={},
  number={},
  pages={3074-3082},
  doi={10.1109/WACV51458.2022.00313}}

@article{yuhao_transfer_learning,
  title = {Sky image-based solar forecasting using deep learning with heterogeneous multi-location data},
  journal = {Appl. Energy},
  volume = {369},
  pages = {123467},
  year = {2024},
  doi = {10.1016/j.apenergy.2024.123467},
  author = {Nie, Y. and Paletta, Q. and Scott, A. and Pomares, L.M. and Arbod, G. and Sgouridis, S. and Lasenby, J. and Brandt, A.},
}

@article{talha2019,
  title={A Deep Learning Approach to Solar-Irradiance Forecasting in Sky-Videos},
  author={Talha Ahmad Siddiqui and Samarth Bharadwaj and Shivkumar Kalyanaraman},
  journal={2019 IEEE Winter Conference on Applications of Computer Vision (WACV)},
  year={2019},
  pages={2166-2174},
}

@article{ineichen1,
author = {Ineichen, Pierre and Perez, Richard},
year = {2002},
month = {09},
pages = {151-157},
title = {A new airmass independent formulation for the Linke turbidity coefficient},
volume = {73},
journal = {Solar Energy},
doi = {10.1016/S0038-092X(02)00045-2}
}

@techreport{clearsky1,
  author       = {Stein, Joshua S and Hansen, Clifford W and Reno, Matthew J},
  title        = {Global horizontal irradiance clear sky models : implementation and analysis.},
  institution  = {Sandia National Laboratories (SNL), Albuquerque, NM, and Livermore, CA (United States)},
  doi          = {10.2172/1039404},
  url          = {https://www.osti.gov/biblio/1039404},
  place        = {United States},
  year         = {2012},
  month        = {03}}

@inproceedings{resnet,
  title={Deep residual learning for image recognition},
  author={He, Kaiming and Zhang, Xiangyu and Ren, Shaoqing and Sun, Jian},
  booktitle={Proceedings of the IEEE conference on computer vision and pattern recognition},
  pages={770--778},
  year={2016}
}

@article{google_vit,
  title={An image is worth 16x16 words: Transformers for image recognition at scale},
  author={Dosovitskiy, Alexey},
  journal={arXiv preprint arXiv:2010.11929},
  year={2020}
}

@article{dinov2,
  title={Dinov2: Learning robust visual features without supervision},
  author={Oquab, Maxime and Darcet, Timoth{\'e}e and Moutakanni, Th{\'e}o and Vo, Huy and Szafraniec, Marc and Khalidov, Vasil and Fernandez, Pierre and Haziza, Daniel and Massa, Francisco and El-Nouby, Alaaeldin and others},
  journal={arXiv preprint arXiv:2304.07193},
  year={2023}
}

@INPROCEEDINGS{uses_freezing_for_tl,
  author={Zhou, S. and Zhou, L. and Mao, M. and Xi, X.},
  booktitle={2020 IEEE Int. Conf. on Big Data and Smart Computing (BigComp)}, 
  title={Transfer Learning for PV Power Forecasting with LSTM Neural Network}, 
  year={2020},
  pages={125-132},
  doi={10.1109/BigComp48618.2020.00-87}
}

@article{transfer_learning_strategies_comparison,
author = {Sarmas, Elissaios and Dimitropoulos, Nikos and Marinakis, Vangelis and Mylona, Zoi and Doukas, H.},
year = {2022},
month = {08},
pages = {},
title = {Transfer learning strategies for solar power forecasting under data scarcity},
volume = {12},
journal = {Scientific Reports},
doi = {10.1038/s41598-022-18516-x}
}

@misc{tsi_dataset,
  author = {Andreas, A. and Stoffel, T.},
  title = {NREL Solar Radiation Research Laboratory (SRRL): Baseline Measurement System (BMS); Golden, Colorado (Data)},
  year = {1981},
  note = {NREL Report No. DA-5500-56488},
  doi = {10.5439/1052221},
}

@article{skippd_dataset,
title = {SKIPP’D: A SKy Images and Photovoltaic Power Generation Dataset for short-term solar forecasting},
journal = {Solar Energy},
volume = {255},
pages = {171-179},
year = {2023},
issn = {0038-092X},
doi = {https://doi.org/10.1016/j.solener.2023.03.043},
author = {Yuhao Nie and Xiatong Li and Andea Scott and Yuchi Sun and Vignesh Venugopal and Adam Brandt}
}

@misc{vits_gt_cnns1,
      title={A Comprehensive Study of Vision Transformers on Dense Prediction Tasks}, 
      author={Kishaan Jeeveswaran and Senthilkumar Kathiresan and Arnav Varma and Omar Magdy and Bahram Zonooz and Elahe Arani},
      year={2022},
      eprint={2201.08683},
      archivePrefix={arXiv},
      primaryClass={cs.CV},
}

@article{prior_methods_site_specific,
author = {Hasan, Ali},
year = {2023},
month = {06},
pages = {},
title = {Predicting Solar Irradiance at Several Time Horizons Using Machine Learning Algorithms}
}

@article{site_specific_trained_transfer_learning,
  title = {Solar irradiance time series forecasting: Influence of transfer learning and clustering},
  journal = {Appl. Energy},
  volume = {365},
  pages = {123215},
  year = {2024},
  doi = {10.1016/j.apenergy.2024.123215},
  author = {Despotovic, M. and Voyant, C. and Garcia-Gutierrez, L. and Almorox, J. and Notton, G.},
}

@article{vaswani2017attention,
  title={Attention is all you need},
  author={Vaswani, A},
  journal={Advances in Neural Information Processing Systems},
  year={2017}
}

@article {nowcasting_defn,
      author = "Jared A. Lee and Sue Ellen Haupt and Pedro A. Jiménez and Matthew A. Rogers and Steven D. Miller and Tyler C. McCandless",
      title = "Solar Irradiance Nowcasting Case Studies near Sacramento",
      journal = "Journal of Applied Meteorology and Climatology",
      year = "2017",
      publisher = "American Meteorological Society",
      address = "Boston MA, USA",
      volume = "56",
      number = "1",
      doi = "10.1175/JAMC-D-16-0183.1",
      pages=      "85 - 108",
}

@article{nowcasting_satellite,
title = {Improved ECMWF forecasts of direct normal irradiance: A tool for better operational strategies in concentrating solar power plants},
journal = {Renewable Energy},
volume = {163},
pages = {755-771},
year = {2021},
issn = {0960-1481},
doi = {https://doi.org/10.1016/j.renene.2020.08.140},
author = {Francis M. Lopes and Ricardo Conceição and Hugo G. Silva and Rui Salgado and Manuel Collares-Pereira}
}

@inproceedings{forecasting_defn,
author = {Remund, Jan and Müller, Stefan},
year = {2012},
month = {08},
pages = {},
title = {SOLAR FORECAST SURVEY RESULTS},
doi = {10.13140/2.1.3826.3681}
}

@article{ghi_pv_linear,
author = {Vilanova, Alba and Kim, Bo-Young and Kim, Chang and Kim, Hyun-Goo},
year = {2020},
month = {02},
pages = {781},
title = {Linear-Gompertz Model-Based Regression of Photovoltaic Power Generation by Satellite Imagery-Based Solar Irradiance},
volume = {13},
journal = {Energies},
doi = {10.3390/en13040781}
}

@article{pv_output, title={Investigation of the Effect Temperature on Photovoltaic (PV) Panel Output Performance}, volume={6}, DOI={10.18517/ijaseit.6.5.938}, number={5}, journal={International Journal on Advanced Science, Engineering and Information Technology}, author={Razak, Amelia and Irwan, Y.M and Leow, W.Z. and Irwanto, M and Safwati, I. and Zhafarina, M.}, year={2016}, month={Oct.}, pages={682–688} }

@article{grid_management,
title = {Seasonal challenges for a California renewable- energy-driven grid},
journal = {iScience},
volume = {25},
number = {1},
pages = {103577},
year = {2022},
issn = {2589-0042},
doi = {https://doi.org/10.1016/j.isci.2021.103577},
author = {Mahmoud Y. Abido and Zabir Mahmud and Pedro Andrés Sánchez-Pérez and Sarah R. Kurtz}
}

@article{object_detection_foundation_models,
  title={Open World Object Detection in the Era of Foundation Models},
  author={Zohar, Orr and Lozano, Alejandro and Goel, Shelly and Yeung, Serena and Wang, Kuan-Chieh},
  journal={arXiv preprint arXiv:2312.05745},
  year={2023}
}

@article{hendrikx2024_short_term_forecast,
  title = {All sky imaging-based short-term solar irradiance forecasting with LSTM networks},
  journal = {Solar Energy},
  volume = {272},
  pages = {112463},
  year = {2024},
  doi = {10.1016/j.solener.2024.112463},
  author = {Hendrikx, N.Y. and Barhmi, K. and Visser, L.R. and {de Bruin}, T.A. and Pó, M. and Salah, A.A. and {van Sark}, W.G.J.H.M.},
}

@article{paletta_transfer_learning,
  title = {Improving cross-site generalisability of vision-based solar forecasting with physics-informed transfer learning},
  journal = {Energy Conv. Manag.},
  volume = {309},
  pages = {118398},
  year = {2024},
  doi = {10.1016/j.enconman.2024.118398},
  author = {Paletta, Q. and Nie, Y. and Saint-Drenan, Y.-M. and {Le Saux}, B.},
}

@misc{efficientnet,
      title={EfficientNet: Rethinking Model Scaling for Convolutional Neural Networks}, 
      author={Mingxing Tan and Quoc V. Le},
      year={2020},
      eprint={1905.11946},
      archivePrefix={arXiv},
      primaryClass={cs.LG},
}

@misc{convnext,
      title={ConvNeXt V2: Co-designing and Scaling ConvNets with Masked Autoencoders}, 
      author={Sanghyun Woo and Shoubhik Debnath and Ronghang Hu and Xinlei Chen and Zhuang Liu and In So Kweon and Saining Xie},
      year={2023},
      eprint={2301.00808},
      archivePrefix={arXiv},
      primaryClass={cs.CV},
}

@misc{aimv2,
      title={Multimodal Autoregressive Pre-training of Large Vision Encoders}, 
      author={Enrico Fini and Mustafa Shukor and Xiujun Li and Philipp Dufter and Michal Klein and David Haldimann and Sai Aitharaju and Victor Guilherme Turrisi da Costa and Louis Béthune and Zhe Gan and Alexander T Toshev and Marcin Eichner and Moin Nabi and Yinfei Yang and Joshua M. Susskind and Alaaeldin El-Nouby},
      year={2024},
      eprint={2411.14402},
      archivePrefix={arXiv},
      primaryClass={cs.CV},
}

\clearpage
\appendix
\appendix
\section{Ablation Studies}
\label{sec:ablation_studies}

\subsection{Impact of the Future Covariates}
We investigate the contribution of the future covariates on SPIRIT's predictive capacity, by removing them and constraining the model to rely exclusively on historical feature vectors. As demonstrated in Table~\ref{tab:future_covariates_ablation}, this modification induces performance degradation across all forecasting horizons. Error increases of 0.62, 1.01, 1.46, and 2.21 percentage points as the prediction window extends from 1-hour to 4-hour forecasts reveal a clear monotonic trend. This progressive deterioration in performance indicates that future covariates provide increasingly valuable information as the prediction horizon extends. The pronounced impact on longer-range forecasts demonstrates that future covariates become particularly crucial when predicting under heightened uncertainty, where historical patterns alone prove insufficient for accurate extrapolation. Incorporation of future covariates enables the architecture to establish more sophisticated temporal dependencies by providing complementary contextual information that augments the historical patterns encoded in past latent representations, thus enhancing the model's ability to capture complex dynamics in the underlying time series. More details are in Section~\ref{detailed_results_with_std_errors}.

\begin{table}[h!]
  \centering
  \setlength{\tabcolsep}{4pt}
  \begin{tabular*}{\columnwidth}{@{\extracolsep{\fill}}lcccc@{}}

    \toprule
    \textbf{Model Configuration} & \multicolumn{4}{c}{\textbf{Forecast}} \\
    \cmidrule(lr){2-5}
    & \textbf{+1hr} & \textbf{+2hr} & \textbf{+3hr} & \textbf{+4hr} \\
    \midrule
    With Future Covariates & \textbf{18.96} & \textbf{21.77} & \textbf{25.46} & \textbf{29.89} \\
    Without Future Covariates & 19.58 & 22.78 & 26.92 & 32.10 \\
    \bottomrule
  \end{tabular*}
  \caption{Ablation study comparing the forecasting performance of SPIRIT with and without future covariates across different time horizons. We train on TSI 2020 and evaluate on TSI 2021 using nMAP error, where lower values indicate better performance.}
  \label{tab:future_covariates_ablation}
\end{table}

\subsection{Investigating Different Vision Encoders}

We evaluate SPIRIT's performance with diverse vision architectures, including CNN-based (ResNet, EfficientNet, ConvNextv2~\cite{convnext}) and attention-based approaches (DINOv2, Google ViT~\cite{google_vit}, Apple AIMv2~\cite{aimv2}). Results in Table \ref{tab:encoders_ablation} show that transformer architectures yield superior performance compared to traditional convolutional approaches, corroborating findings that self-attention mechanisms better capture the global image features \cite{vits_gt_cnns1} critical for solar forecasting. Notably, ConvNeXtv2 matches ViTs in performance, owing to its novel FCMAE framework and GRN mechanism~\cite{convnext}. The architectural modularity of SPIRIT facilitates seamless integration of such emerging models, conferring substantial advantages for sustained operational deployment.

\begin{table}[h!]
  \footnotesize
  \setlength{\tabcolsep}{4pt}
  \begin{tabular*}{\columnwidth}{@{\extracolsep{\fill}}lcrrrrr@{}}
    \toprule
    \textbf{Model} & \textbf{Params} & \multicolumn{1}{c}{\textbf{Nowcast}} & \multicolumn{4}{c}{\textbf{Forecast}} \\
    \cmidrule(lr){4-7}
    & & &  \textbf{+1hr} & \textbf{+2hr} & \textbf{+3hr} & \textbf{+4hr} \\
    \midrule
    AIMv2 & 2.72B & 9.04 & 20.89 & 23.90 & 27.22 & 32.71 \\
    DINOv2 & 1.14B & 9.74 & 21.22 & 23.56 & 27.93 & 33.13 \\
    Google & 632M & \textbf{9.04} & \textbf{18.96} & \textbf{21.77} & 25.46 & 29.89 \\
    \midrule
    ConvNeXtv2 & 88.7M & 9.92 & 19.73 & 22.09 & \textbf{25.17} & \textbf{29.56} \\
    EfficientNet-B7 & 66M & 13.34 & 25.20 & 27.25 & 29.58 & 34.27 \\
    ResNet-152 & 60.3M & 10.50 & 24.56 & 27.82 & 31.23 & 35.85 \\
    \bottomrule
  \end{tabular*}
  \caption{We use the largest variants of diverse CNN and transformer-based vision encoders to explore their influence on SPIRIT's nowcasting and forecasting performance, training on TSI 2020 and testing on TSI 2021, measured by nMAP error.
  }
  \label{tab:encoders_ablation}
\end{table}

\subsection{Foundation Model Size}

Table~\ref{tab:encoder_sizes_ablation} presents an analysis of how the size of the foundation model influences the performance of our nowcasting and forecasting architectures. Although increasing model size has traditionally been linked to performance gains, we observe that beyond a threshold, further scaling yields diminishing returns or even performance degradation as with DINOv2. For transformer-based architectures, models in the moderate parameter range outperform the largest variants, while in the ResNet family, the smallest variant demonstrates superior forecasting performance, whereas the second smallest attains peak nowcasting results.

These patterns suggest that the impact of model scaling aligns more closely with parameter count than with architectural family distinctions. Specifically, for transformer-based models such as DINO and Google-ViT, performance improves consistently with increasing size up to approximately 600M parameters, beyond which gains plateau or regress. In contrast, ResNets exhibit distinct scaling dynamics that diverge from those of transformer architectures.

\begin{table}[h!]
  \small
  \setlength{\tabcolsep}{4pt} 
  \begin{tabular*}{\columnwidth}{@{\extracolsep{\fill}}lcrrrrr@{}}
    \toprule
    \textbf{Model} & \textbf{Params} & \multicolumn{1}{c}{\textbf{Nowcast}} & \multicolumn{4}{c}{\textbf{Forecast}} \\
     \cmidrule(lr){4-7}
    & & & \textbf{+1hr} & \textbf{+2hr} & \textbf{+3hr} & \textbf{+4hr} \\
    \midrule
    \multirow{3}{*}{Google ViT} & 86M & 9.14 & 21.92 & 24.07 & 28.73 & 34.50 \\
    & 304M & 9.45 & 19.58 & 21.95 & 25.54 & 30.60 \\
    & 632M & \textbf{9.04} & \textbf{18.96} & \textbf{21.77} & \textbf{25.46} & \textbf{29.89} \\
    \midrule
    \multirow{3}{*}{DINOv2} & 22.1M & 10.13 & 20.08 &22.53 & 25.69 & 29.86 \\
    & 86.6M & \textbf{9.56} & 19.52 & 21.95 & \textbf{25.00} & \textbf{29.57} \\
    & 304M & 10.04 & \textbf{19.32} & \textbf{21.74} & 25.29 & 30.04 \\
    & 1.14B & 9.74 & 21.22 & 23.56 & 27.93 & 33.13 \\
    \midrule
    \multirow{4}{*}{ResNet} & 11.7M & 10.47 & \textbf{22.31} & \textbf{24.28} & \textbf{27.28} & \textbf{32.52} \\
    & 25.6M & \textbf{10.29} & 24.36 & 25.93 & 29.07 & 34.10 \\
    & 44.5M & 10.97 & 24.99 & 26.21 & 30.62 & 34.86 \\
    & 60.3M & 10.50 & 24.56 & 27.82 & 31.23 & 35.85 \\
    \bottomrule
  \end{tabular*}
  \caption{Impact of scaling across Google-ViT, DINOv2, and ResNet model families on nowcasting and forecasting, with training on TSI 2020 and testing on TSI 2021, and nMAP as the error metric. Results are highlighted for the best performer in each family.
  }
  \label{tab:encoder_sizes_ablation}
\end{table}

\subsection{Contribution of Auxiliary and Physics Features}

To quantify the individual and combined contributions of different feature modalities, we conduct a systematic ablation study examining the impact of image embeddings, physics-based features, and auxiliary spatiotemporal variables on nowcasting performance. As presented in Table \ref{tab:nowcasting_ablation}, the results reveal several critical insights regarding feature importance and complementarity. The baseline configuration using only image embeddings achieves a test nMAP of 12.45\%, establishing the fundamental predictive capability of visual information. The incorporation of either physics features (9.48\% nMAP) or auxiliary features (9.55\% nMAP) individually provides substantial performance improvements, with both modalities contributing nearly equivalent error reductions. Crucially, the combination of physics and auxiliary features without image embeddings yields severely degraded performance (34.23\% nMAP), conclusively demonstrating that visual information serves as the primary predictive signal, while physics-based and auxiliary features provide essential complementary enhancements rather than standalone predictive power. The optimal configuration integrating all three feature types achieves the lowest error at 9.04\% nMAP, validating our feature choice and the importance of our multimodal approach in capturing the complex dynamics inherent in this task.

\begin{table}[htbp]
  \centering
  \setlength{\tabcolsep}{8pt}
  \renewcommand{\arraystretch}{1.2}
  \begin{tabular}{l c}
    \hline
    \textbf{Setup} & \textbf{nMAP (\%)} \\
    \hline
    Only image embeddings              & 12.45 \\
    Embeddings + Physics               & 9.48 \\
    Embeddings + Auxiliary             & 9.55 \\
    Physics + Auxiliary                & 34.23 \\
    Embeddings + Physics + Auxiliary   & \textbf{9.04} \\
    \hline
  \end{tabular}
  \caption{
  Ablation on the contribution of image embeddings, physics features, and auxiliary features for nowcasting. Results are reported as test nMAP (\%, lower is better). The model is trained on TSI-2020 and tested on TSI-2021.
  }
  \label{tab:nowcasting_ablation}
\end{table}

\subsection{Variation of Regressors for Nowcasting}

To establish the optimal modeling approach for nowcasting, we conducted a comprehensive evaluation of different regression models in our system while maintaining consistent input features across all models. As demonstrated in Table \ref{tab:nowcasting_regressors}, XGBoost achieves superior performance with a test nMAP of 9.04\%, substantially outperforming alternative approaches. This performance advantage can be attributed to XGBoost's inherent strengths in handling heterogeneous tabular data, where its gradient boosting framework effectively captures complex interactions between auxiliary meteorological features, physics-based variables, and high-dimensional image embeddings. The ensemble nature of XGBoost enables it to leverage the complementary information from these diverse feature modalities more effectively than traditional neural architectures in this specific nowcasting context. 

\begin{table}[htbp]
  \centering
  \setlength{\tabcolsep}{8pt}
  \renewcommand{\arraystretch}{1.2}
  \begin{tabular}{l c}
    \hline
    \textbf{Regressor} & \textbf{nMAP (\%)} \\
    \hline
    MLP        & 19.74 \\
    KNN Regressor        & 16.51 \\
    Random Forest Regressor & 15.66 \\
    XGBoost    & \textbf{9.04} \\
    \hline
  \end{tabular}
  \caption{
  Comparison of different regressors for nowcasting using the same input features. Results are reported as test nMAP (\%, lower is better). XGBoost performs the best, indicating its suitability for this task. The models are trained on TSI 2020 and evaluated on TSI 2021.
  }
  \label{tab:nowcasting_regressors}
\end{table}

\section{Dataset Details}
\label{sec:appendix_dataset_details}

\begin{table*}
  \centering
  \renewcommand{\arraystretch}{1.2}
  \begin{tabularx}{\textwidth}{X X X X}
    \hline
    \textbf{Attribute} & \textbf{TSI880 Dataset} & \textbf{ASI16 Dataset} & \textbf{SKIPP'D Dataset} \\ 
    \hline
    Location & Golden, Colorado, USA & Golden, Colorado, USA & Stanford, California, USA \\ 
    Data Type & Sky images \& Irradiance data & Sky images \& Irradiance data & Sky images \& PV power output \\ 
    Data Frequency & 10-minutes & 10-minutes & 1-minute \\ 
    Image Resolution & 288x352 & 1536x1536 & 64x64 \\ 
    Camera Model & Aero-Laser TSI-880 & EKO ASI-16 & Hikvision DS-2CD6362F-IV \\ 
    Number of Samples / Year & 24,948 & 25,107 & 121,125 \\ 
    \hline
  \end{tabularx}
  \caption{A Comparative Overview of the TSI880, ASI16, and SKIPP'D Datasets: Key Attributes Including Geographical Location, Data Provided, Image Resolution, Collection Frequency, and Annual Sample Size}
  \label{tab:dataset_comparison}
\end{table*}

\subsection{Overview of Datasets}
\label{subsec:appendix_overview_of_datasets}
\textbf{TSI880 Dataset:}
The TSI880 dataset is collected from the NREL Solar Radiation Research Laboratory in Golden, Colorado. The camera captures an image every 10 minutes from 7:50 to 16:40 daily, providing raw sky images along with corresponding global horizontal irradiance values. Additionally, the dataset includes auxiliary information such as air temperature, relative humidity, azimuth angle, and zenith angle.

\textbf{ASI16 Dataset:}
The ASI16 dataset is also sourced from the Solar Radiation Research Laboratory in Golden, Colorado, but it differs in that the camera setup captures images at a higher resolution. Similar to the TSI880 dataset, it provides global horizontal irradiance values and auxiliary data including azimuth angle, zenith angle, air temperature, relative humidity, and average wind speed.

\textbf{SKIPP'D Dataset:}
The SKIPP'D dataset consists of raw sky images and photovoltaic (PV) power output data collected from Stanford University, California, USA. Images are captured every minute with a resolution of 64×64 pixels, emphasizing finer temporal granularity at the expense of lower image resolution.

\subsection{Robustness to Low-Quality Visual Inputs}
The SKIPP'D dataset helps us assess the model's robustness under challenging visual conditions. With a resolution of just 64×64 pixels, the SKIPP'D images are significantly lower in quality than those found in most practical deployments, where sky imagers typically produce higher-resolution outputs. By evaluating on SKIPP'D, we test whether our model can still perform reliably in scenarios where image quality is degraded due to hardware limitations, compression, or transmission artifacts. This setup provides a realistic stress test, ensuring that our model is not only accurate under ideal conditions, but also resilient in more constrained, real-world environments.

\begin{figure}[t]
    \centering
\includegraphics[width=200pt]{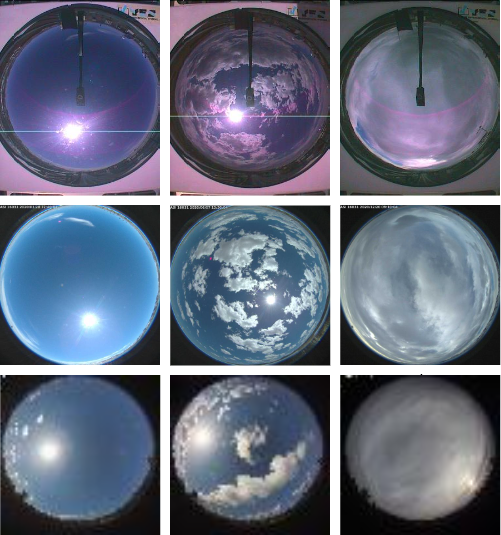}
    
    \caption{Examples of sunny, partly cloudy, and overcast conditions, captured by different sky cameras, are shown from left to right, across the three datasets: TSI, ASI, and SKIPP'D, displayed from top to bottom.
    \\
    }
    \label{fig:dataset_grid}
\end{figure}

\subsection{Temporal Consistency in Forecasting}
\label{subsec:appendix_temporal_consistency_in_forecasting}
Valid samples for forecasting are formed such that all the data points from time steps \( 1 \) to \( T \), and their corresponding forecast intervals \( T + \tau_1, T + \tau_2, \dots, T + \tau_H \), fall within the same day. This is an essential requirement because the predictions for future intervals rely on the assumption that both historical and forecast data belong to the same day. Using data from the current day to predict values for the following day is not a valid forecasting approach, as the discontinuity between days renders such predictions unreliable. Any samples that violate this condition are considered invalid and are excluded from training or evaluation.

\section{Clear Sky Global Horizontal Irradiance}
\label{sec:appendix_clear_sky_global_horizontal_irradiance}
Clear Sky Global Horizontal Irradiance (GHI) is the solar irradiance received on a horizontal surface under cloud-free conditions. Most of the time, it serves as an upper bound for the actual GHI at a given location and time.

Clear Sky GHI plays a key role in solar forecasting by serving as a baseline for estimating how much clouds reduce solar irradiance. By comparing actual irradiance with Clear Sky GHI, we can get an estimate of the impact of cloud cover, which helps in enhancing short-term predictions, and improving the accuracy of forecasting models.

Given the latitude and longitude of a location, the clear sky values can be estimated for any timestamp. This becomes very useful in solar forecasting, as this value would give a reference of how much the prediction needs to be.

Clear Sky GHI is computed using mathematical models incorporating solar position, atmospheric transmittance, and radiative transfer principles. A common approach is the Ineichen-Perez model :

\begin{equation}
    GHI_{\text{clear}} = I_0 \cdot \tau \cdot \cos(\theta_z)
\end{equation}

where \( I_0 \) is the extraterrestrial irradiance (W/m²), \( \tau \) is the atmospheric transmittance factor, \( \theta_z \) is the solar zenith angle.









\section{Fine-tuning Detailed Results}
\label{sec:appendix_fine-tuning_detailed_results}
\subsection{Nowcasting}
\label{subsec:appendix_nowcasting}
To understand the impact of fine-tuning duration and the training size, we conducted a series of experiments by varying the amount of training data used for fine-tuning, by using subsets of the data consisting of 1, 2, 3, and 4 weeks.

Our results show that even with only one week of training data at a new location, the fine-tuned model performs remarkably well. Furthermore, in all experimental configurations, our model significantly outperforms the baseline.

Detailed results for these experiments are presented in Tables~\ref{tab:oneweek_nowcast}, \ref{tab:twoweek_nowcast}, \ref{tab:threeweek_nowcast}, and~\ref{tab:fourweek_nowcast}.

\begin{table}[htbp]
  \centering
  \renewcommand{\arraystretch}{1.2}
  \begin{tabular}{@{}c c c c@{}}
    \hline
    \textbf{Trained on} & \textbf{Finetuned on} & \textbf{SPIRIT} & \textbf{\citeauthor{wacv2022}} \\
    \hline
    \multirow{3}{*}{TSI} & ASI & \textbf{20.23}  & 52.01 \\
                         & SKIPP'D & \textbf{29.89}  & 63.82 \\
    \cline{1-4}
    \multirow{3}{*}{ASI} & TSI & \textbf{14.99}  & 27.98 \\
                         & SKIPP'D & \textbf{27.51}  & 40.92 \\
    \hline
  \end{tabular}
  \caption{Nowcasting performance with one week of training.}
  \label{tab:oneweek_nowcast}
\end{table}

\begin{table}[htbp]
  \centering
  \renewcommand{\arraystretch}{1.2}
  \begin{tabular}{@{}c c c c@{}}
    \hline
    \textbf{Trained on} & \textbf{Finetuned on} & \textbf{SPIRIT} & \textbf{\citeauthor{wacv2022}} \\
    \hline
    \multirow{2}{*}{TSI} & ASI & \textbf{18.96}  & 51.45 \\
                          & SKIPP'D & \textbf{29.07}  & 62.91 \\
    \cline{1-4}
    \multirow{2}{*}{ASI} & TSI & \textbf{14.91}  & 27.71 \\
                          & SKIPP'D & \textbf{26.41}  & 40.25 \\
    \hline
  \end{tabular}
  \caption{
  Nowcasting Performance with 2 weeks training
  }
  \label{tab:twoweek_nowcast}
\end{table}

\begin{table}[htbp]
  \centering
  \renewcommand{\arraystretch}{1.2}
  \begin{tabular}{@{}c c c c@{}}
    \hline
    \textbf{Trained on} & \textbf{Finetuned on} & \textbf{SPIRIT} & \textbf{\citeauthor{wacv2022}} \\
    \hline
    \multirow{2}{*}{TSI} & ASI & \textbf{16.52}  & 50.38 \\
                          & SKIPP'D & \textbf{27.42}  & 62.05 \\
    \cline{1-4}
    \multirow{2}{*}{ASI} & TSI & \textbf{14.59}  & 27.53 \\
                          & SKIPP'D & \textbf{25.68}  & 39.89 \\
    \hline
  \end{tabular}
  \caption{
  Nowcasting Performance with 3 weeks training
  }
  \label{tab:threeweek_nowcast}
\end{table}

\begin{table}[htbp]
  \centering
  \renewcommand{\arraystretch}{1.2}
  \begin{tabular}{@{}c c c c@{}}
    \hline
    \textbf{Trained on} & \textbf{Finetuned on} & \textbf{SPIRIT} & \textbf{\citeauthor{wacv2022}} \\
    \hline
    \multirow{2}{*}{TSI} & ASI & \textbf{15.63}  & 50.01 \\
                          & SKIPP'D & \textbf{26.51}  & 61.17 \\
    \cline{1-4}
    \multirow{2}{*}{ASI} & TSI & \textbf{14.12}  & 27.28 \\
                          & SKIPP'D & \textbf{24.32}  & 39.43 \\
    \hline
  \end{tabular}
  \caption{
  Nowcasting Performance with 4 weeks training
  }
  \label{tab:fourweek_nowcast}
\end{table}

\subsection{Forecasting}
\label{subsec:appendix_forecasting}
\begin{table}[htbp]
  \centering
  \setlength{\tabcolsep}{2pt}
  \renewcommand{\arraystretch}{1.2} 
  \begin{tabular}{c c c c c}
    \hline
    \textbf{Interval} & \textbf{Trained on} & \textbf{Tested on} & \textbf{SPIRIT} & \textbf{\citeauthor{wacv2022}} \\
    \hline
    \multirow{4}{*}{1hr} & \multirow{2}{*}{TSI} & ASI & \textbf{31.15} & 33.86 \\
                          & & SKIPP'D & \textbf{32.35} & 38.24 \\
                          \cline{2-5}
                          & \multirow{2}{*}{ASI} & TSI & \textbf{24.47} & 36.18 \\
                          & & SKIPP'D & \textbf{27.00} & 30.48 \\
    \hline
    \multirow{4}{*}{2hr} & \multirow{2}{*}{TSI} & ASI & \textbf{32.70} & 36.44 \\
                          & & SKIPP'D & \textbf{29.41} & 39.06 \\
                          \cline{2-5}
                          & \multirow{2}{*}{ASI} & TSI & \textbf{25.93} & 36.71 \\
                          & & SKIPP'D & \textbf{25.96} & 33.55 \\
    \hline
    \multirow{4}{*}{3hr} & \multirow{2}{*}{TSI} & ASI & \textbf{34.41} & 38.24 \\
                          & & SKIPP'D & \textbf{31.53} & 39.84 \\
                          \cline{2-5}
                          & \multirow{2}{*}{ASI} & TSI & \textbf{30.45} & 41.46 \\
                          & & SKIPP'D & \textbf{30.03} & 39.76 \\
    \hline
    \multirow{4}{*}{4hr} & \multirow{2}{*}{TSI} & ASI & \textbf{38.19} & 43.76 \\
                          & & SKIPP'D & \textbf{36.83} & 41.76 \\
                          \cline{2-5}
                          & \multirow{2}{*}{ASI} & TSI & \textbf{36.44} & 45.89 \\
                          & & SKIPP'D & \textbf{36.84} & 44.16 \\
    \hline
  \end{tabular}
  \caption{
  Forecasting Performance with 2 weeks of training.
  }
  \label{tab:twoweek_forecast}
\end{table}

\begin{table}[htbp]
  \centering
  \setlength{\tabcolsep}{2pt}
  \renewcommand{\arraystretch}{1.2} 
  \begin{tabular}{c c c c c}
    \hline
    \textbf{Interval} & \textbf{Trained on} & \textbf{Tested on} & \textbf{SPIRIT} & \textbf{\citeauthor{wacv2022}} \\
    \hline
    \multirow{4}{*}{1hr} & \multirow{2}{*}{TSI} & ASI & \textbf{22.17} & 29.03 \\
                          & & SKIPP'D & \textbf{32.44} & 39.82 \\
                          \cline{2-5}
                          & \multirow{2}{*}{ASI} & TSI & \textbf{27.65} & 35.77 \\
                          & & SKIPP'D & \textbf{26.54} & 30.70 \\
    \hline
    \multirow{4}{*}{2hr} & \multirow{2}{*}{TSI} & ASI & \textbf{25.13} & 32.69 \\
                          & & SKIPP'D & \textbf{29.56} & 40.21 \\
                          \cline{2-5}
                          & \multirow{2}{*}{ASI} & TSI & \textbf{31.06} & 36.62 \\
                          & & SKIPP'D & \textbf{25.53} & 33.63 \\
    \hline
    \multirow{4}{*}{3hr} & \multirow{2}{*}{TSI} & ASI & \textbf{30.12} & 38.64 \\
                          & & SKIPP'D & \textbf{31.79} & 40.18 \\
                          \cline{2-5}
                          & \multirow{2}{*}{ASI} & TSI & \textbf{34.47} & 38.76 \\
                          & & SKIPP'D & \textbf{29.73} & 39.70 \\
    \hline
    \multirow{4}{*}{4hr} & \multirow{2}{*}{TSI} & ASI & \textbf{36.14} & 41.92 \\
                          & & SKIPP'D & \textbf{37.24} & 41.31 \\
                          \cline{2-5}
                          & \multirow{2}{*}{ASI} & TSI & \textbf{39.72} & 40.02 \\
                          & & SKIPP'D & \textbf{36.67} & 44.16 \\
    \hline
  \end{tabular}
  \caption{
  Forecasting Performance with 4 weeks of training.
  }
  \label{tab:fourweek_forecast}
\end{table}

\begin{table}[htbp]
  \centering
  \setlength{\tabcolsep}{2pt}
  \renewcommand{\arraystretch}{1.2} 
  \begin{tabular}{c c c c c}
    \hline
    \textbf{Interval} & \textbf{Trained on} & \textbf{Tested on} & \textbf{SPIRIT} & \textbf{\citeauthor{wacv2022}} \\
    \hline
    \multirow{4}{*}{1hr} & \multirow{2}{*}{TSI} & ASI & \textbf{22.62} & 32.45 \\
                          & & SKIPP'D & \textbf{33.56} & 36.94 \\
                          \cline{2-5}
                          & \multirow{2}{*}{ASI} & TSI & \textbf{26.38} & 35.70 \\
                          & & SKIPP'D & \textbf{26.61} & 31.25 \\
    \hline
    \multirow{4}{*}{2hr} & \multirow{2}{*}{TSI} & ASI & \textbf{25.15} & 33.58 \\
                          & & SKIPP'D & \textbf{30.65} & 38.06 \\
                          \cline{2-5}
                          & \multirow{2}{*}{ASI} & TSI & \textbf{26.68} & 35.26 \\
                          & & SKIPP'D & \textbf{25.30} & 33.95 \\
    \hline
    \multirow{4}{*}{3hr} & \multirow{2}{*}{TSI} & ASI & \textbf{28.66} & 35.57 \\
                          & & SKIPP'D & \textbf{32.64} & 39.29 \\
                          \cline{2-5}
                          & \multirow{2}{*}{ASI} & TSI & \textbf{29.81} & 36.44 \\
                          & & SKIPP'D & \textbf{29.25} & 39.85 \\
    \hline
    \multirow{4}{*}{4hr} & \multirow{2}{*}{TSI} & ASI & \textbf{34.76} & 39.41 \\
                          & & SKIPP'D & \textbf{37.80} & 41.63 \\
                          \cline{2-5}
                          & \multirow{2}{*}{ASI} & TSI & \textbf{34.97} & 38.23 \\
                          & & SKIPP'D & \textbf{36.23} & 44.25 \\
    \hline
  \end{tabular}
  \caption{
  Forecasting Performance with 8 weeks of training.
  }
  \label{tab:eightweek_forecast}
\end{table}

\begin{table}[htbp]
  \centering
  \setlength{\tabcolsep}{2pt}
  \renewcommand{\arraystretch}{1.2} 
  \begin{tabular}{c c c c c}
    \hline
    \textbf{Interval} & \textbf{Trained on} & \textbf{Tested on} & \textbf{SPIRIT} & \textbf{\citeauthor{wacv2022}} \\
    \hline
    \multirow{4}{*}{1hr} & \multirow{2}{*}{TSI} & ASI & \textbf{22.03} & 34.63 \\
                          & & SKIPP'D & \textbf{33.76} & 37.35 \\
                          \cline{2-5}
                          & \multirow{2}{*}{ASI} & TSI & \textbf{24.87} & 35.24 \\
                          & & SKIPP'D & \textbf{28.28} & 31.20 \\
    \hline
    \multirow{4}{*}{2hr} & \multirow{2}{*}{TSI} & ASI & \textbf{24.95} & 35.81 \\
                          & & SKIPP'D & \textbf{30.38} & 38.16 \\
                          \cline{2-5}
                          & \multirow{2}{*}{ASI} & TSI & \textbf{27.42} & 35.31 \\
                          & & SKIPP'D & \textbf{26.17} & 35.01 \\
    \hline
    \multirow{4}{*}{3hr} & \multirow{2}{*}{TSI} & ASI & \textbf{29.86} & 38.02 \\
                          & & SKIPP'D & \textbf{31.80} & 39.12 \\
                          \cline{2-5}
                          & \multirow{2}{*}{ASI} & TSI & \textbf{30.04} & 36.61 \\
                          & & SKIPP'D & \textbf{29.61} & 41.13 \\
    \hline
    \multirow{4}{*}{4hr} & \multirow{2}{*}{TSI} & ASI & \textbf{34.37} & 41.27 \\
                          & & SKIPP'D & \textbf{36.60} & 41.34 \\
                          \cline{2-5}
                          & \multirow{2}{*}{ASI} & TSI & \textbf{35.71} & 38.67 \\
                          & & SKIPP'D & \textbf{36.16} & 45.28 \\
    \hline
  \end{tabular}
  \caption{
  Forecasting Performance with 12 weeks of training.
  }
  \label{tab:twelveweek_forecast}
\end{table}

\begin{table}[htbp]
  \centering
  \setlength{\tabcolsep}{2pt}
  \renewcommand{\arraystretch}{1.2} 
  \begin{tabular}{c c c c c}
    \hline
    \textbf{Interval} & \textbf{Trained on} & \textbf{Tested on} & \textbf{SPIRIT} & \textbf{\citeauthor{wacv2022}} \\
    \hline
    \multirow{4}{*}{1hr} & \multirow{2}{*}{TSI} & ASI & \textbf{22.76} & 28.97 \\
                          & & SKIPP'D & \textbf{33.12} & 36.93 \\
                          \cline{2-5}
                          & \multirow{2}{*}{ASI} & TSI & \textbf{23.33} & 35.07 \\
                          & & SKIPP'D & \textbf{25.74} & 31.01 \\
    \hline
    \multirow{4}{*}{2hr} & \multirow{2}{*}{TSI} & ASI & \textbf{25.30} & 31.55 \\
                          & & SKIPP'D & \textbf{30.75} & 38.18 \\
                          \cline{2-5}
                          & \multirow{2}{*}{ASI} & TSI & \textbf{27.48} & 36.57 \\
                          & & SKIPP'D & \textbf{25.83} & 32.22 \\
    \hline
    \multirow{4}{*}{3hr} & \multirow{2}{*}{TSI} & ASI & \textbf{28.86} & 36.28 \\
                          & & SKIPP'D & \textbf{33.10} & 39.83 \\
                          \cline{2-5}
                          & \multirow{2}{*}{ASI} & TSI & \textbf{31.92} & 39.46 \\
                          & & SKIPP'D & \textbf{31.04} & 37.69 \\
    \hline
    \multirow{4}{*}{4hr} & \multirow{2}{*}{TSI} & ASI & \textbf{33.99} & 41.36 \\
                          & & SKIPP'D & \textbf{38.20} & 42.66 \\
                          \cline{2-5}
                          & \multirow{2}{*}{ASI} & TSI & \textbf{37.50} & 42.14 \\
                          & & SKIPP'D & \textbf{38.25} & 42.46 \\
    \hline
  \end{tabular}
  \caption{
  Forecasting Performance with 16 weeks of training.
  }
  \label{tab:sixteenweek_forecast}
\end{table}

We conducted a series of experiments to assess the impact of training data size on model performance during fine-tuning. We utilized training splits of 2, 4, 8, 12, and 16 weeks of data at the new site. For each training duration, we performed experiments with different random splits of the corresponding number of weeks and reported the results accordingly.

The results are presented in Tables~\ref{tab:twoweek_forecast}, \ref{tab:fourweek_forecast}, \ref{tab:eightweek_forecast}, \ref{tab:twelveweek_forecast}, and \ref{tab:sixteenweek_forecast}. The figure for forecasting performance upon finetuning (Figure 3 in the main paper) was constructed by systematically aggregating the results from our fine-tuning experiments, encapsulating the performance trends observed across different training durations. By leveraging visualization techniques, the figure provides a holistic representation of how the model adapts as more site-specific data becomes available. It effectively summarizes variations in performance across different random splits of finetuning data and across different sets of source and target datasets. This covers training on TSI, and testing on ASI and SKIPP'D, as well as training on ASI, and testing on TSI and SKIPP'D. This comprehensive validation methodology substantiates SPIRIT's efficacy across diverse deployment conditions during transfer learning.

We employed 95\% confidence intervals for all experiments, spanning diverse transfer learning settings and random sampling of the fine-tuning data. To rigorously compare our method with the baseline across different weekly intervals, we applied a paired t-test at a significance level of 0.001 (i.e., less than a 0.1\% chance of incorrectly rejecting the null hypothesis). In every instance, the observed p-values fell below this threshold, demonstrating that SPIRIT achieves statistically significant performance improvements over the baseline.

\subsection{Frozen Vision Encoder in Fine-tuning Strategy}

We perform selective fine-tuning on the regressor and time-series transformer while keeping the vision encoder frozen. This is a deliberate methodological choice made to validate our core hypothesis that generalized vision encoders outperform site-specific ones used in prior work. Unfreezing the encoder during fine-tuning would introduce location-specific biases and domain-specific adaptations, which would contradict our fundamental objective of demonstrating the superiority of foundation model-based visual representations for cross-domain solar irradiance prediction. By maintaining the frozen encoder, we ensure that the observed performance improvements are attributable to the inherent generalizability of pre-trained vision features rather than location and camera configuration specific adaptations, thereby providing a rigorous evaluation of our approach's transferability across diverse deployment scenarios.

\section{Zero-Shot Transfer Learning Results for More Foundation Models}

To comprehensively evaluate the generalization capabilities of our proposed system, we extend our zero-shot transfer learning analysis beyond the Google ViT Large model to encompass a diverse range of foundation models, including multiple variants of the DINO family (spanning from 22.1M to 1.1B parameters), ConvNeXtV2, and EfficientNet architectures. As demonstrated in Table \ref{tab:zero_shot_all_models}, our findings reveal a consistent and compelling pattern across all evaluated models and transfer scenarios.

While our system demonstrates comparable or incrementally superior performance to the baseline within the same geographical location with the same camera setup, the true strength of our approach emerges when evaluated across different deployment contexts. Remarkably, every variation of our system, regardless of the underlying vision encoder architecture, consistently outperforms the baseline method when transferring to new locations and camera configurations. This consistent superiority manifests across both cross-camera transfer scenarios, where models trained on TSI2020 are evaluated on ASI2021, and cross-location with cross-task generalization scenarios, exemplified by the transfer from TSI2020 to SKIPP'D2017.

The robustness of this performance advantage across architecturally diverse models spanning traditional CNN-based approaches like ConvNeXtV2 and EfficientNet, as well as various scales of vision transformer architectures provides compelling evidence for our central hypothesis that generalized vision encoders possess superior transfer learning capabilities for atmospheric forecasting applications. This systematic outperformance across different encoder types, parameter scales, and training paradigms underscores the fundamental strength of our system design in leveraging pre-trained visual representations for cross-domain atmospheric prediction tasks. The consistency of these improvements across diverse architectural choices validates that our enhanced system design, rather than specific model selection, drives the observed generalization benefits.

\begin{table*}[htbp]
\centering
\renewcommand{\arraystretch}{1.2}
\setlength{\tabcolsep}{6pt}
\begin{tabular}{c c c c c c}
\hline
\textbf{Test Dataset} & \textbf{Model} & \textbf{1-Hour} & \textbf{2-Hour} & \textbf{3-Hour} & \textbf{4-Hour} \\
\hline
\multirow{7}{*}{TSI}
& {\citeauthor{wacv2022}} & 19.96 & 22.64 & 26.30 & 31.58 \\
\cline{2-6}
& DINO-304M & \textbf{19.32} & \textbf{21.74} & 25.29 & 30.04 \\
& DINO-86.6M & 19.52 & 21.95 & \textbf{25.00} & \textbf{29.57} \\
& DINO-22.1M & 20.08 & 22.53 & 25.69 & 29.86 \\
& DINO-1.1B & 21.22 & 23.56 & 27.93 & 33.13 \\
& ConvNeXtV2 & 19.73 & 22.09 & 25.17 & 29.56 \\
& EfficientNet & 25.20 & 27.25 & 29.58 & 34.27 \\
\hline
\multirow{7}{*}{ASI}
& {\citeauthor{wacv2022}} & 35.74 & 37.60 & 37.77 & 39.58 \\
\cline{2-6}
& DINO-304M & \textbf{30.11} & \textbf{31.86} & 34.52 & 37.91 \\
& DINO-86.6M & 30.30 & 32.01 & \textbf{34.49} & 38.21 \\
& DINO-22.1M & 31.22 & 33.16 & 35.60 & 39.06 \\
& DINO-1.1B & 32.41 & 34.93 & 36.55 & 40.12 \\
& ConvNeXtV2 & 30.42 & 32.19 & 34.84 & \textbf{38.91} \\
& EfficientNet & 35.06 & 37.15 & 39.11 & 41.94 \\
\hline
\multirow{7}{*}{SKIPP'D}
& {\citeauthor{wacv2022}} & 38.88 & 43.81 & 47.36 & 51.78 \\
\cline{2-6}
& DINO-304M & 33.20 & 30.95 & 32.62 & 35.71 \\
& DINO-86.6M & \textbf{33.15} & \textbf{30.79} & 32.89 & 35.97 \\
& DINO-22.1M & 34.05 & 31.37 & 33.21 & 37.11 \\
& DINO-1.1B & 34.95 & 32.75 & 34.50 & 38.78 \\
& ConvNeXtV2 & 33.71 & 31.16 & \textbf{32.97} & \textbf{36.12} \\
& EfficientNet & 36.51 & 34.67 & 35.98 & 40.29 \\
\hline
\end{tabular}
\caption{
Zero-Shot Forecasting Results for Multiple Vision Foundation Models (used as encoders in our pipeline) trained on TSI2020 and tested on TSI2021 (TSI), ASI2021 (ASI), and SKIPP'D2017 (SKIPP'D). All values represent forecasting error (lower is better). Results for the baseline model by \citeauthor{wacv2022} are also included for comparison. Bold indicates the best value per column within each test dataset.
}
\label{tab:zero_shot_all_models}
\end{table*}

\section{Implementation Details}
Our implementation incorporates azimuth and zenith angles as auxiliary inputs, from which we derive the physics-based features utilizing the pvlib library: clear sky global horizontal irradiance (GHI), clear sky diffuse horizontal irradiance (DHI), and clear sky direct normal irradiance (DNI) values. We then employ these values to obtain the effective irradiance on the panel. The auxiliary and physics-based parameters serve dual functions: they generate future covariates and also provide statistical context to the time series encoder regarding past timestamps. We use clear sky GHI values as residual inputs to the final MLP layer which contributes favorably to the model's overall predictive capabilities. Comprehensive hyperparameter optimization was conducted across all reported experimental configurations to ensure methodological consistency and reproducibility of results.


\subsection{Hyperparameter Configuration}
For nowcasting, we use XGBoost on a 1297-dimensional feature vector (comprising 1280 ViT features and 17 physics/auxiliary features). The hyperparameters are: \texttt{max\_depth} = 7, \texttt{learning\_rate} = 0.021, \texttt{n\_estimators} = 1386, \texttt{subsample} = 0.653, \texttt{colsample\_bytree} = 0.888, \texttt{gamma} = 0.002, \texttt{lambda} = 1.744, and \texttt{early\_stopping\_rounds} = 200. 

For forecasting, we use a 6-layer Transformer with 8 attention heads and a hidden dimension of 1104, along with learnable positional encodings. The regressor consists of 10 MLP blocks with residual connections. Clear-sky GHI is added as a residual input before the final layer. The model is trained using stochastic gradient descent (SGD) with a learning rate of 0.00032, momentum of 0.9, and a warm-up followed by a cosine annealing learning rate schedule. Hyperparameter tuning is performed using Optuna.

\subsection{Normalization}
We perform z-score normalization on both the global horizontal irradiance (GHI) and photovoltaic (PV) output signals, standardizing them to have zero mean and unit variance. This normalization facilitates stable training and allows the model to directly learn the underlying relationship between sky conditions and energy output without requiring additional bias correction. To ensure continuity in time-series inputs, missing image frames or sensor readings are handled by interpolating via the average of their nearest neighboring values, which preserves temporal coherence and minimizes disruptions during model training or inference.

\subsection{Real-Time Deployment Efficiency and Performance}
SPIRIT demonstrates excellent suitability for real-world deployment, with an average inference time of just 0.5378 seconds (±0.0046) on a GPU with 24GB VRAM and 16 CPU cores. This low latency ensures that predictions are generated well within the available time window between consecutive image frames. In our evaluation datasets, SKIPP'D has the highest temporal resolution with a 1-minute gap between frames, while TSI and ASI, which are representative of most operational sky image datasets, feature 10-minute intervals. Given this context, our model's inference speed is significantly faster than required, confirming that latency is not a bottleneck for real-time solar forecasting applications.

\section{Other Baselines}
\label{sec:baseline_justification}

\subsection{Limitations of Statistical Models}
Classical time series forecasting methods such as ARIMA and Vector AutoRegression (VAR) are frequently used for structured prediction tasks due to their simplicity and interpretability. However, these techniques operate solely on temporal sequences and are fundamentally limited in their ability to capture the spatiotemporal complexity inherent in sky imagery. These models lack the visual context necessary to model phenomena such as cloud occlusion, movement, and changes, which are critical for accurate solar irradiance forecasting.

\subsection{ARIMA: Univariate Modeling of GHI}
The ARIMA model combines autoregressive (AR) and moving average (MA) components with differencing to handle non-stationary time series. For a time series $y_t$, the ARIMA$(p,d,q)$ model is expressed as $\phi(L)(1-L)^d y_t = \theta(L)\epsilon_t$, where $\phi(L) = 1 - \phi_1 L - \phi_2 L^2 - \cdots - \phi_p L^p$ is the autoregressive polynomial, $\theta(L) = 1 + \theta_1 L + \theta_2 L^2 + \cdots + \theta_q L^q$ is the moving average polynomial, $L$ is the lag operator, and $\epsilon_t$ is white noise. For our ARIMA$(2,0,2)$ model applied to GHI, this becomes $y_t = \phi_1 y_{t-1} + \phi_2 y_{t-2} + \epsilon_t + \theta_1 \epsilon_{t-1} + \theta_2 \epsilon_{t-2}$, where $y_t$ represents the GHI value at time $t$.

As a statistical baseline, we employ the Autoregressive Integrated Moving Average (ARIMA) model on a univariate time series of Global Horizontal Irradiance (GHI). The order $(2, 0, 2)$ was selected based on Akaike Information Criterion (AIC) and confirmed to be appropriate after verifying stationarity using the Augmented Dickey-Fuller (ADF) test. The differencing order $d=0$ was chosen since the time series was found to be stationary.

The ARIMA model was trained on the TSI 2020 dataset and evaluated on both the TSI 2021 and ASI 2021 datasets. The forecasting horizon was set to 4 hours, and only the preceding 1 hour of data (equivalent to 6 frames with 10-minute intervals) was used as context, in alignment with SPIRIT's input and output windows.
The ARIMA model yielded a Mean Absolute Error (MAE) of 168.94, a Root Mean Square Error (RMSE) of 195.13, and a negative $R^2$ score of -0.6180, with a normalized Mean Absolute Percentage (nMAP) of 55.51\%. These results were consistent across both the TSI 2021 and ASI 2021 datasets. This is expected, as the ARIMA model operates purely on GHI values and does not utilize any image-based input. Since the datasets share the same auxiliary irradiance data and originate from the same NREL facility in Colorado, the only difference in camera configuration has no effect on this statistical model's performance.

\subsection{VAR: Multivariate Time Series Forecasting}
The Vector AutoRegression (VAR) model extends univariate autoregressive modeling to multivariate time series, where each variable is regressed on its own lagged values and the lagged values of all other variables in the system. For a VAR$(p)$ model with $k$ variables, the system is represented as $\mathbf{y}_t = \mathbf{c} + \mathbf{A}_1 \mathbf{y}_{t-1} + \mathbf{A}_2 \mathbf{y}_{t-2} + \cdots + \mathbf{A}_p \mathbf{y}_{t-p} + \mathbf{e}_t$, where $\mathbf{y}_t$ is a $k \times 1$ vector of endogenous variables at time $t$, $\mathbf{c}$ is a $k \times 1$ vector of constants, $\mathbf{A}_i$ are $k \times k$ coefficient matrices for $i = 1, 2, \ldots, p$, and $\mathbf{e}_t$ is a $k \times 1$ vector of white noise error terms. In our implementation, $\mathbf{y}_t$ contains clear-sky GHI, clear-sky DNI, actual GHI, and physics-based auxiliary variables, enabling the model to capture interdependencies between these solar irradiance components.

We also evaluate a Vector AutoRegression (VAR) model trained on multivariate time series composed of clear-sky GHI, clear-sky Direct Normal Irradiance (DNI), actual GHI, and physics-based auxiliary variables. To maintain consistency with the image-based baselines, we limit the context window to the previous 6 time steps (1 hour), setting \texttt{maxlags}=6.
The VAR model achieved an MAE of 147.25, RMSE of 174.99, an $R^2$ value of -0.3013, and a normalized MAP of 48.38\% on both the TSI 2021 and ASI 2021 datasets. This uniformity in performance is due to the shared irradiance and auxiliary variables used as input, which are identical across the two datasets. As with ARIMA, the VAR model does not utilize visual data, and therefore its forecasting capability is unaffected by the differing camera configurations of TSI and ASI.

\subsection{The Limitations of Purely Statistical Approaches}
Despite being well-established in traditional forecasting pipelines, autoregressive statistical models, whether univariate (ARIMA) or multivariate (VAR), fall significantly short when tasked with predicting solar irradiance. Their inability to process visual inputs makes them fundamentally incapable of modeling the intricate spatial patterns and dynamic cloud formations captured in sky images. Both ARIMA and VAR produce high error rates and even negative $R^2$ values, highlighting their limited generalization capabilities. In contrast, vision-based models like ours exhibit strong predictive accuracy even under location shifts. These findings underscore the necessity of incorporating spatiotemporal visual cues for solar irradiance forecasting.

\subsection{On CNN-LSTM Baselines}

We benchmark our approach against the current state-of-the-art method~\cite{wacv2022}, which has demonstrated superior performance on publicly accessible datasets and established the performance ceiling for solar forecasting tasks. Vision transformer architectures have fundamentally superseded CNN-LSTM approaches across forecasting domains, with~\cite{wacv2022} which outperforms traditional convolutional methods~\cite{talha2019}. Our experimental design prioritizes meaningful evaluation against the strongest available competitor, ensuring that any performance gains represent genuine technological advancement in solar forecasting capabilities.

The established dominance of transformer-based architectures in computer vision forecasting tasks makes comparisons with CNN-LSTM baselines redundant as they have been systematically outperformed. Our focused comparison allows for a clear analysis of architectural innovations in solar forecasting, without aiming to reestablish existing performance hierarchies.

\section{Detailed Results with Standard Errors}
\label{detailed_results_with_std_errors}
Table~\ref{tab:detailed_results_std_error} reports the detailed forecasting performance of our model, including standard errors computed across multiple random trials. The consistently small standard errors across all forecast horizons demonstrate the robustness and stability of our approach. In contrast, Table~\ref{tab:without_future_covariates} presents results obtained when the model is trained and evaluated without future covariates, showing markedly degraded accuracy. 
Importantly, the performance gaps between the two settings are substantially larger than the associated standard errors, indicating that the observed improvements are significant. This underscores the critical role of future covariates in enhancing predictive accuracy for longer forecast horizons.

\begin{table}[htbp]
  \centering
  \setlength{\tabcolsep}{8pt}
  \renewcommand{\arraystretch}{1.2}
  \begin{tabular}{c c}
    \hline
    \textbf{Forecast Horizon} & \textbf{Mean ± Std. Error} \\
    \hline
    Nowcasting      & 9.080 ± 0.00526 \\
    1-hour forecast & 19.136 ± 0.092 \\
    2-hour forecast & 21.779 ± 0.101 \\
    3-hour forecast & 25.317 ± 0.170 \\
    4-hour forecast & 29.780 ± 0.259 \\
    \hline
  \end{tabular}
  \caption{
  Detailed forecasting performance of our model with mean values and standard errors across random trials. The results show consistent improvements with low variance for nowcasting and future forecasts.
  }
  \label{tab:detailed_results_std_error}
\end{table}

\begin{table}[htbp]
  \centering
  \setlength{\tabcolsep}{8pt}
  \renewcommand{\arraystretch}{1.2}
  \begin{tabular}{c c}
    \hline
    \textbf{Forecast Horizon} & \textbf{Mean ± Std. Error} \\
    \hline
    1-hour forecast & 19.688 ± 0.085 \\
    2-hour forecast & 22.858 ± 0.088 \\
    3-hour forecast & 26.800 ± 0.083 \\
    4-hour forecast & 31.890 ± 0.095 \\
    \hline
  \end{tabular}
  \caption{
  Forecasting performance of the model \textit{without future covariates}, showing degraded accuracy across horizons. The gap between these results and the main model highlights the importance of future covariates.
  }
  \label{tab:without_future_covariates}
\end{table}

\section{Comprehensive Analysis of Evaluation Metrics}

The evaluation of solar forecasting models requires careful consideration of appropriate metrics that accurately reflect model performance in real-world deployment scenarios. We employ multiple evaluation metrics to provide a comprehensive assessment of our proposed methodology. The normalized Mean Absolute Percentage error (nMAP) is defined as:
\begin{equation}
\text{nMAP} = \frac{1}{N} \sum_{i=1}^{N} \frac{|y_i - \hat{y}_i|}{\frac{1}{N}\sum_{i=1}^{N} y_i} \times 100
\end{equation}
Additionally, we evaluate performance using Mean Absolute Error (MAE):
\begin{equation}
\text{MAE} = \frac{1}{N} \sum_{i=1}^{N} |y_i - \hat{y}_i|
\end{equation}
Root Mean Square Error (RMSE):
\begin{equation}
\text{RMSE} = \sqrt{\frac{1}{N} \sum_{i=1}^{N} (y_i - \hat{y}_i)^2}
\end{equation}
and coefficient of determination ($R^2$):
\begin{equation}
R^2 = 1 - \frac{\sum_{i=1}^{N} (y_i - \hat{y}_i)^2}{\sum_{i=1}^{N} (y_i - \bar{y})^2}
\end{equation}
where $y_i$ represents the actual value, $\hat{y}_i$ represents the predicted value for the $i$-th sample, and $\bar{y}$ is the mean of actual values, with $i \in \{1, \ldots, N\}$. The comprehensive results across all these metrics are presented in Table~\ref{tab:comprehensive_metrics}.

\begin{table}[h!]
  \centering
  \setlength{\tabcolsep}{5pt}
  \renewcommand{\arraystretch}{1.2}
  \begin{tabular}{cccccc}
    \hline
    \textbf{Interval} & \textbf{Tested on} & \textbf{nMAP} & \textbf{RMSE} & \textbf{R\textsuperscript{2}} & \textbf{MAE} \\
    \hline
    \multirow{3}{*}{1 hr}
      & TSI 2021   & 18.96 & 176.60 & 0.56 & 106.63 \\
      & ASI 2021   & 29.99 & 200.05 & 0.50 & 129.11 \\
      & SKIPP’D    & 32.93 & 205.11 & 0.42 & 134.19 \\
    \hline
    \multirow{3}{*}{2 hr}
      & TSI 2021   & 21.77 & 195.07 & 0.49 & 120.80 \\
      & ASI 2021   & 31.71 & 203.01 & 0.43 & 132.06 \\
      & SKIPP’D    & 29.01 & 198.53 & 0.47 & 127.54 \\
    \hline
    \multirow{3}{*}{3 hr}
      & TSI 2021   & 25.46 & 201.73 & 0.53 & 129.66 \\
      & ASI 2021   & 34.41 & 207.59 & 0.45 & 136.87 \\
      & SKIPP’D    & 30.26 & 200.56 & 0.52 & 129.58 \\
    \hline
    \multirow{3}{*}{4 hr}
      & TSI 2021   & 29.89 & 199.73 & 0.60 & 128.77 \\
      & ASI 2021   & 38.00 & 214.55 & 0.58 & 143.26 \\
      & SKIPP’D    & 34.63 & 207.96 & 0.61 & 137.21 \\
    \hline
  \end{tabular}
  \caption{
    SPIRIT’s generalization performance across datasets and time intervals. The model is trained on TSI 2020 and evaluated on TSI 2021, ASI 2021, and SKIPP’D using four forecasting horizons (1 hr–4 hr). We report four metrics: nMAP, RMSE, R\textsuperscript{2}, and MAE.
  }
  \label{tab:comprehensive_metrics}
\end{table}

We primarily focus on nMAP as our principal evaluation metric due to its established status as the standard metric in solar forecasting research. This standardization facilitates meaningful comparisons across different methodologies and ensures our contributions align with established benchmarks in the field. The practical significance of nMAP improvements extends far beyond academic metrics, as even marginal improvements translate to substantial economic and environmental benefits when deployed across large-scale solar installations. These improvements compound across extensive solar deployments, potentially saving millions in operational costs while simultaneously reducing carbon emissions through enhanced grid stability and reduced reliance on fossil fuel backup systems.

The normalized nature of nMAP ensures that our improvements maintain consistent significance across diverse geographic conditions and varying solar irradiance patterns. This geographic invariance is crucial for developing universally applicable forecasting models that can be deployed across different climatic zones and solar resource conditions without requiring region-specific recalibration of performance expectations. While we conduct comprehensive multi-metric evaluation during model development and validation as shown in Table~\ref{tab:comprehensive_metrics}, we deliberately emphasize nMAP as the decisive evaluation metric throughout our analysis to maintain consistency with established solar forecasting standards and provide clear, interpretable results.



\section{Other Details}

\textbf{Longer forecasts:}
SPIRIT supports longer-horizon forecasts. However, our problem setting focuses on short-term forecasting, critical for grid operators who need high-resolution predictions for real-time dispatch decisions and load balancing. Beyond 4 hours,the focus shifts away from grid control, and accuracy deteriorates due to atmospheric randomness and clouds moving beyond the camera's field-of-view, which is why longer-term forecasting relies on numerical weather prediction(NWP) and satellites, representing a fundamentally different forecasting paradigm.

\textbf{Zero-Shot Transfer and Historical Data Requirements:}

To clarify our methodology's data requirements, our approach eliminates the need for historical data at new deployment sites. New solar installations typically require 5-10 years of site-specific data before providing reliable forecasts, creating a major bottleneck in renewable energy expansion and delaying the deployment of effective forecasting systems. SPIRIT addresses this fundamental challenge by leveraging datasets from established sites to achieve zero-shot transfer to new locations, enabling immediate deployment of forecasting systems without any data collection period. This capability is crucial for accelerating solar energy adoption, as it eliminates the traditional waiting period required for accumulating sufficient historical data at each new installation site. While our model does utilize historical data during training from established sites, the key innovation lies in transferring this learned knowledge to new sites without requiring any site-specific historical data collection. 


\end{document}